\documentclass[twocolumn,amsmath,amssymb,aps,pra,]{revtex4-1}
\usepackage{geometry}
\usepackage{graphicx}
\usepackage{dcolumn}
\usepackage{bm}
\usepackage{lipsum}
\usepackage{color}
\usepackage{textcomp}

\newif\ifcmnt
\cmnttrue

\ifcmnt
    \providecommand{\aucmnt}[1]{#1}
\else
    \providecommand{\aucmnt}[1]{}
\fi

\begin{document}
	
	\title[]{Superconducting optoelectronic circuits for neuromorphic computing}
	
	\author{Jeffrey M. Shainline}
	\email{jeffrey.shainline@nist.gov}
	\affiliation{National Institute of Standards and Technology, 325 Broadway, Boulder, CO, USA}
	
	\author{Sonia M. Buckley}
	\affiliation{National Institute of Standards and Technology, 325 Broadway, Boulder, CO, USA}	
	
	\author{Richard P. Mirin}
	\affiliation{National Institute of Standards and Technology, 325 Broadway, Boulder, CO, USA}
		
	\author{Sae Woo Nam}
	\affiliation{National Institute of Standards and Technology, 325 Broadway, Boulder, CO, USA}

	\date{\today}
	
	\begin{abstract}
	Neural networks have proven effective for solving many difficult computational problems. Implementing complex neural networks in software is very computationally expensive. To explore the limits of information processing, it will be necessary to implement new hardware platforms with large numbers of neurons, each with a large number of connections to other neurons. Here we propose a hybrid semiconductor-superconductor hardware platform for the implementation of neural networks and large-scale neuromorphic computing. The platform combines semiconducting few-photon light-emitting diodes with superconducting-nanowire single-photon detectors to behave as spiking neurons. These processing units are connected via a network of optical waveguides, and variable weights of connection can be implemented using several approaches. The use of light as a signaling mechanism overcomes fanout and parasitic constraints on electrical signals while simultaneously introducing physical degrees of freedom which can be employed for computation. The use of supercurrents achieves the low power density necessary to scale to systems with enormous entropy. The proposed processing units can operate at speeds of at least $20$ MHz with fully asynchronous activity, light-speed-limited latency, and power densities on the order of 1 mW/cm$^2$ for neurons with 700 connections operating at full speed at 2 K. The processing units achieve an energy efficiency of $\approx 20$ aJ per synapse event. By leveraging multilayer photonics with deposited waveguides and superconductors with feature sizes $>$ 100 nm, this approach could scale to systems with massive interconnectivity and complexity for advanced computing as well as explorations of information processing capacity in systems with an enormous number of information-bearing microstates.
	\end{abstract}
	
	
	\maketitle

	
	\section{\label{sec:intro}Introduction}	
	Many foundational concepts in information theory and computing were developed beginning in the 1930s and 1940s through the work of Turing \cite{tu1936}, von Neumann \cite{ne1945}, Shannon \cite{sh1948}, and others. Given the variety of proposed approaches to computing, it is somewhat surprising that the current landscape of computing technologies exclusively uses the von Neumann architecture. There has long been an interest in the relationship between information, computation, and cognition \cite{tu1950,ne1958}. Computing architectures drawing inspiration from biological neural systems have been considered for decades \cite{me1990}, but investigation of novel architectures is only now becoming urgent as we reach the end of Moore's law scaling. The recent surge in deep learning and neural networks, marked by advances in hardware \cite{chsl2012,paha2014,mear2014}, applications \cite{sihu2016}, and theory \cite{rapo2016,pola2016,lite2016} has increased our understanding of the importance of such systems for solving complex problems.
	
	Lin and Tegmark have recently argued \cite{lite2016} that the physics of our universe is conducive to representation by neural networks. While there are an infinite number of possible functions a network may try to approximate, only a very limited subset will be of interest in our physical world. Additionally, it has been shown mathematically \cite{rapo2016,pola2016} that the ability of a neural network to accurately represent different kinds of functions (the expressivity of the network) scales as $k^{mn}$, where $m$ is the dimension of the input, $n$ is the number of hidden layers, and $k$ is the number of nodes in each layer. This insight informs us that we can improve a network's ability to represent a broad range of functions both by increasing its width ($k$) and depth ($n$). Further, since the total information capacity of a computing system is proportional to the entropy, which scales with the number of distinct states which can be addressed by the system \cite{ma1999}, computing systems based on complex interconnected networks, such as biological neural systems, offer extraordinary computational power. 
	
	To further maximize the information processing capacity of such a system, it is desirable to fully utilize the time domain. For resilience to noise as well as temporally encoded information \cite{pasc1999,haah2015}, signal communication via pulses, or spikes, is most advantageous, and such spike-encoded information is most powerful when many connections are established between processing units \cite{haah2015}. All of these findings taken together inform us that to implement neural networks most effectively in hardware, we should develop systems with a large total number of processing units, a large number of connections between units, and pulse-based communication.
		
\begin{figure} 
		\centerline{\includegraphics[width=7.0cm]{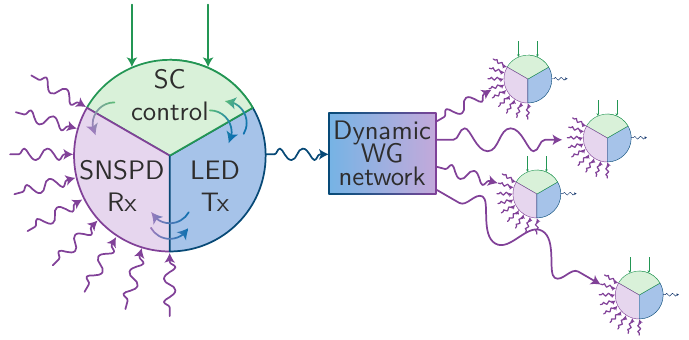}}
		\caption{\label{fig:cartoon} Schematic representation of the proposed device concept. SC = superconducting; SNSPD = superconducting-nanowire single-photon detector; Rx = receive; Tx = transmit; WG = waveguide.}
	\end{figure}

	Much like the von Neumann architecture has dominated modern computing, the hardware of silicon microelectronics has been similarly preeminent. It is possible that the ideal hardware platform for the next generation of computer architectures will also look very different. We make two conjectures which lead us to the hardware platform presented here. The first is that photons, based on their non-interacting, bosonic nature, will prove advantageous over electrons for achieving spike-based communication over networks with a large number of connections between nodes. That is to say, photonic fanout will overcome limitations of electronic fanout. The second conjecture is that superconducting circuits will enable lower power densities than semiconducting circuits, thereby leading to systems with a larger number of processing units and greater total complexity. In conceiving of a hardware platform integrating photonic with superconducting devices, we find a feasible route to highly scaled, multi-physical systems with extraordinary potential for computing complexity and experiments in information physics. A schematic representation of the concept is shown in Fig. \ref{fig:cartoon}.
	
	The optoelectronic hardware platform is based on waveguide-integrated semiconductor light emitters working with superconducting detectors and electronics to implement weighted, directed networks \cite{bagu2007}. Optical signals between neurons are communicated through reconfigurable nanophotonic waveguides. Utilization of light-emitting semiconductors allows efficient access to photonic degrees of freedom (frequency, polarization, mode index, intensity, statistics, and coherence) which achieve complex functionality analogous to chemical signaling in biological organisms, and possibly with information processing capabilities far beyond. Light enables massive interconnectivity with no need for time-multiplexing schemes that can limit the event rates of complimentary metal-oxide-semiconductor (CMOS) systems \cite{mear2014,hama2013}. Photonic signals are received and integrated by superconducting single-photon detectors. Firing thresholds and gain are controlled by a dynamic superconducting network, and neuron-generated photonic signals can reconfigure this current-distribution network. By employing superconducting electronics, we can approach zero static power dissipation \cite{kisa2011}, extraordinary device efficiencies, and utilize Josephson junction circuits including single-flux-quantum devices \cite{hias2007,crsc2010,ru2016}.
	
	Within this hardware platform, memory can be implemented via several means. These include temporally fixed synapses achieved with branching waveguides; synaptic weight variation via the actuation of locally suspended waveguides; or through the use of magnetic Josephson junctions \cite{vevi2013} or other magnetic and flux-storage components. The suspended waveguides that we explore in more detail in this work are reconfigurable on a time scale of $1\mu s$. None of these approaches draw power in the steady state.
	
	The combination of efficient faint-light sources and superconducting-nanowire single-photon detectors interacting in an integrated-photonics environment enables neuronal operation with excellent energy efficiency, enormous intra- and inter-chip communication bandwidth, light-speed-limited latency, compact footprint, and relatively simple fabrication. The optoelectronic hardware platform is predicted to achieve 20 aJ/synapse event. By comparison, many CMOS systems are on the order of 20 pJ/synapse event \cite{inli2011,wusa2015,mear2014}.  

	The article is organized as follows. In Sec. \ref{sec:SPON} we present the foundational neuronal optoelectronic circuits and consider each of the requisite constituent components. In Sec. \ref{sec:connectivity} we discuss the coupling of these circuits as well as mechanisms for reconfigurable memory enabling plasticity and learning. In Sec. \ref{sec:apps} we discuss concrete applications of this hardware platform and consider the spatial and power scaling. We conclude with Sec. \ref{sec:discussion}. Details of the device design are presented in the appendices.
								
	\section{\label{sec:SPON}Optoelectronic neuronal circuits}
	Information in neural systems is often referred to as ``spike encoded'' as interconnected neurons transmit information to one another in pulses \cite{daab2001}. An individual neuron (also referred to as a ``processing unit'' or simply ``unit''), will receive pulses from a number of upstream neurons.  The neuron's input/output relation will be nonlinear, and if the integrated upstream signals exceed a certain threshold, the neuron may itself fire a pulse to its downstream connections. In this section we describe superconducting optoelectronic circuits to emulate several biological neural responses. These circuits use integrated LEDs as transmitters with optical detectors as receivers. We next discuss the requirements for detectors and LEDs for this platform, and we motivate our choice from current technologies. Based on these choices, the energy per firing event is calculated.
	
	\subsection{\label{sec:snspd}Detector choice}
	A neuron that uses photonic signals requires both a source of photons and a photon detector. The choice of detector is critical to the design and analysis of the hardware platform. The central aim of this hardware platform is to achieve massive scaling to large numbers of interacting neurons. Therefore, simple waveguide integration, extreme energy efficiency, high yield, and small size are principal concerns. A review and comparison of single-photon detectors can be found in Ref. \cite{eisaman2011}. Of all existing detector options, only those based on superconductors allow single-photon detection in the infrared with zero static power dissipation and single-photon sensitivity to enable operation at the shot-noise limit. Because a system based on superconducting detectors would enable operation in this limit, it would offer a useful platform to test the role of noise in learning and evolution of complex, dynamical systems.
	
	There is an additional energy cost associated with cooling superconducting detectors to cryogenic temperatures necessary for operation. Therefore an alternative is to move away from low-light levels, and use integrated detectors such as Si \cite{Woodward1999,Choi2010,Tanabe2010}, Si defect \cite{Ackert2011,meor2014}, Ge-on-Si \cite{Assefa2010,Assefa2010a,Michel2010} or III-V detectors, either bonded to Si \cite{Brouckaert2007} or on a fully III-V platform \cite{Nagarajan2010}. Such detectors have low signal-to-noise ratio, requiring operation with significantly higher optical powers than if superconducting detectors are employed. While it may be possible to develop neuromorphic technology based on many of these detectors, we have chosen for this article to focus on superconducting nanowire single photon detectors (SNSPDs) due to the high efficiencies ($>$ 90\%) \cite{mave2013}  at wavelengths below the Si bandgap, simple on-chip waveguide integration \cite{spga2011,pesc2012,feka2015,namo2015,saga2015,scgu2016,shbu2016}, compact size, and speed. While operation at cryogenic temperatures imparts a fixed energy cost, the energy cost per operation is significantly decreased by allowing integration with superconducting electronics. Therefore, cryogenic systems are of use in a subset of neuromorphic applications where the required system size is sufficiently large that the savings in chip power outweigh the cryocooling cost. Additionally, low-temperature operation allows the use of certain LED designs that are not possible at room temperature, as will be discussed in Sec. \ref{sec:LED}.
	
	\subsection{\label{sec:parallel}Integrate-and-fire circuit}
\begin{figure} 
		\centerline{\includegraphics[width=7.0cm]{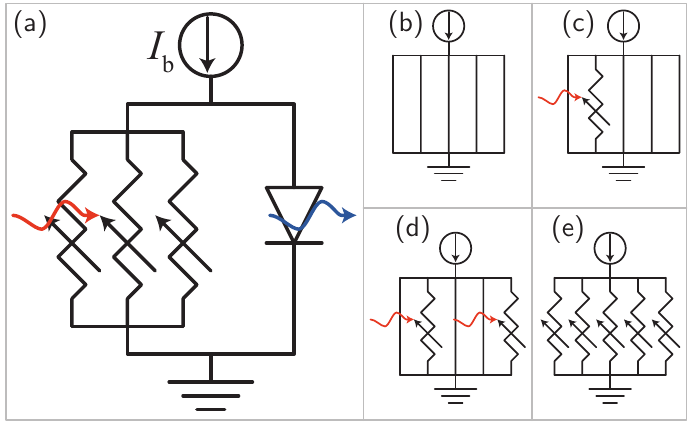}}
		\caption{\label{fig:circuits_PND} (a) PND neuron circuit. (b) A PND with all wires superconducting (c) A PND where one of the wires has been driven normal by absorption of a single photon, redirecting the current through the other four. (d) A PND with two normal wires due to absorption of two photons (e) A PND with all wires driven normal by exceeding the critical current. An LED in parallel with this PND will now receive current, causing a firing event.}
	\end{figure}

    To encode information, the nodes of a neural network must have a nonlinear input-output relationship. In the proposed system, that nonlinearity is achieved via the transition of wires from the superconducting phase to the normal-metal phase. These phase transitions can be induced by absorption of a photon or by exceeding the critical current. A single SNSPD can be designed to fire with close to unity efficiency upon absorbing a single photon. We can think of this as an integrate-and-fire neuron in the limit of a single-photon threshold. In order to obtain an integrate-and-fire response with a threshold photon number larger than one, SNSPDs can be configured in parallel (step response) or series (continuous response). In Fig. \ref{fig:circuits_PND}(a) we show a circuit diagram of the parallel SNSPD array, referred to as a parallel nanowire detector (PND) \cite{dima2008,mabi2009}. One example of an integrate-and-fire circuit is accomplished by placing the PND in parallel with an LED. The thresholding mechanism is explained pictorially in Fig. \ref{fig:circuits_PND}(b)-(e). In the steady state, the PND is superconducting and has zero resistance. The semiconducting LED has finite resistance, and therefore all current from the source $I_\mathrm{b}$ flows through the PND. When a sufficient number of nanowires in the PND has been driven to the normal state by the absorption of photons, the critical current of the array is exceeded, the array becomes resistive, and current is diverted to the LED. This diversion of current and the subsequent production of light via carrier recombination constitutes the firing event. The LED fires with a step response, meaning that the LED output is independent of the exact number of photons absorbed, and only depends on whether or not the threshold has been exceeded. The diversion of current to the LED allows the PND to return to the superconducting state. Once this has occurred, current ceases to flow through the LED, the production of light stops, and the device is reset.
	
	
	The minimum duration of a spike event is determined by the emitter lifetime. The integration time of the neuron can be engineered to be within the range of a few hundred picoseconds up seconds. See Appendix \ref{apx:time} for more detailed discussion of the temporal response of the circuits.
	
	\begin{figure} 
		\centerline{\includegraphics[width=7.0cm]{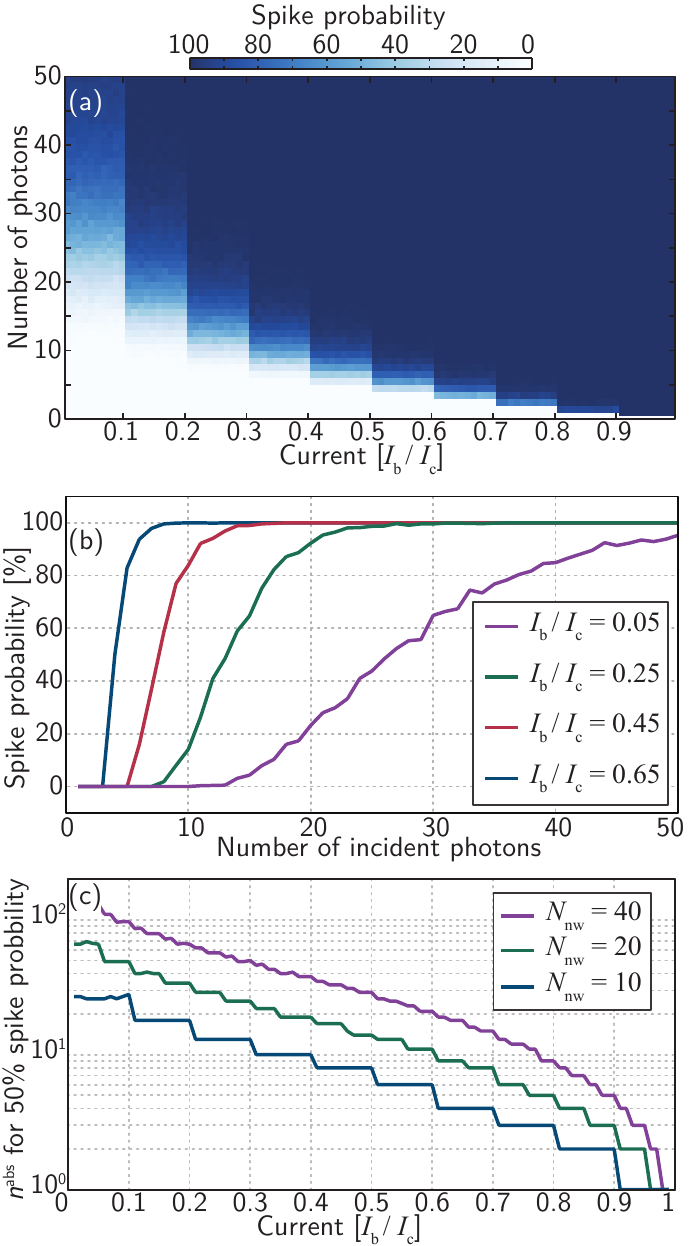}}
		\caption{\label{fig:spikeProbability} Monte-Carlo simulation of spike probability. (a) PND with 10 SNSPDs. (b) The same simulation as (a), but with four traces isolated for clarity. (c) The number of absorbed photons which gives a $50\%$ absorption probability plotted as a function of bias current. Traces for PNDs with 10, 20, and 40 nanowires are shown.}
	\end{figure}
\begin{figure} 
		\centerline{\includegraphics[width=7.0cm]{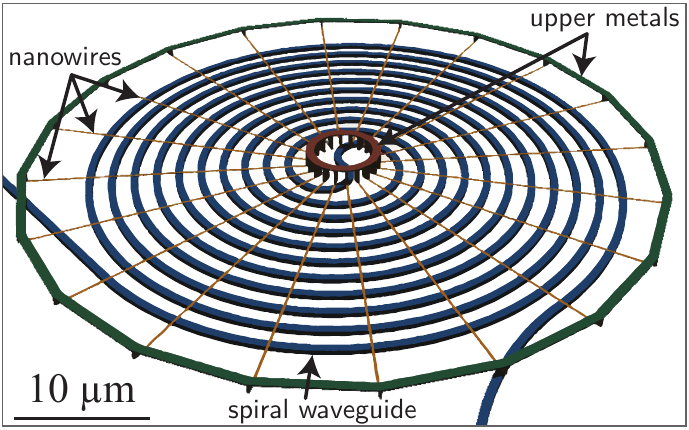}}
		\caption{\label{fig:spiderweb_1} The spiderweb neuron. The scale bar is shown for reference, but significantly more compact implementations of this device could be achieved.}
	\end{figure}

	To model the spike probability of this circuit, we have conducted Monte Carlo simulations of the device. The critical number of absorbed photons, $n_{\mathrm{c}}$, is given by	
	\begin{equation}
	\label{eq:nc}
	n_{\mathrm{c}} = N_{\mathrm{nw}}-\frac{I_{\mathrm{b}}}{i_{\mathrm{c}}},
	\end{equation}
	where $N_{\mathrm{nw}}$ is the number of nanowires in the array, $I_\mathrm{b}$ is the bias current for the entire array, and $i_{\mathrm{c}}$ is the critical current of a single wire. Equation \ref{eq:nc} is derived in Appendix \ref{apx:PNDAppendix}. Although each individual firing event generates the same current pulse across the LED (i.e. a step response), a given number of input photons will only cause the neuron to fire with some probability. This is due to the stochastic nature of the photon absorption events, which is discussed in more detail in Appendix \ref{apx:SNSPDAppendix}. The results of these simulations are shown in Fig. \ref{fig:spikeProbability}. The probability of a spike occurring is plotted as a function of the number of photons incident on the device for various bias currents ranging from 0.01 of the array critical current ($I_{\mathrm{c}}$) to 0.99 $I_{\mathrm{c}}$ in steps of 0.01 $I_{\mathrm{c}}$. In Fig. \ref{fig:spikeProbability}(a) we show the behavior of an array with 10 SNSPDs in parallel. Figure \ref{fig:spikeProbability}(b) shows spike probability versus the number of incident photons for four values of bias current; this data is a subset of that shown in Fig. \ref{fig:spikeProbability}(a), plotted separately to illustrate the shape of the traces. The Monte Carlo simulations which produced these plots are conceptually based on the neuron design of Fig. \ref{fig:spiderweb_1}, and proceed as follows. A given number of photons was assumed to be incident on a PND array. The pulse was assumed to pass each nanowire of the array in sequence. At each pass, a random number between zero and one was generated. If this random number was less than or equal to the assumed absorption probability (1\% in these calculations), the number of photons in the pulse was reduced by one, and the state of that nanowire was set to non-superconducting. The photon pulse was allowed to pass each nanowire of the array 100 times. The number of photons in the pulse which caused Eq. \ref{eq:nc} to be satisfied was recorded for each bias current. The result of 1,000 such simulations was averaged to calculate the probability for spiking to occur. 
	
	In Figs. \ref{fig:spikeProbability}(a) and (b), we observe that by adjusting the bias current we can adjust the shape of the firing function versus photon number. Yet adjusting the bias current cannot tune the threshold with arbitrary accuracy. In Fig. \ref{fig:spikeProbability}(a), it is evident that the spike probability for a PND array with 10 nanowires separates into ten bands. Therefore, to achieve higher photon number differentiation, more wires must be integrated. This point is illustrated in Fig. \ref{fig:spikeProbability}(c). Simulations similar to that of Fig. \ref{fig:spikeProbability}(a) were conducted for PND arrays with 20 and 40 nanowires, and the number of absorbed photons ($n^{\mathrm{abs}}$) for which the spike probability reached 50\% is plotted versus the bias current. This figure further illustrates that the resolution of the PND array is limited by the number of nanowires in the array, resulting in discrete steps in the number of photons required for a spike event as a function of bias current. Because $n_{\mathrm{c}}$ and $N_{\mathrm{nw}}$ in Eq. \ref{eq:nc} are both integers, the floor of the ratio $I_\mathrm{b}/i_{\mathrm{c}}$ is effectively taken, and the utility of the current for setting the threshold is discretized. For the case of $N_{\mathrm{nw}}=40$, the steps become quite small, and the curve is approximately continuous.
	
	The simple model of Fig. \ref{fig:spikeProbability} reveals that the PND array can achieve a high dynamic range in that the threshold can be tuned broadly in hardware by changing the number of wires in the array (from a single nanowire up to potentially thousands) as well as actively during operation by changing the bias current. The state space of the receiver, which scales as $2^{N_{\mathrm{nw}}}$, can be made quite large in the regime where thousands of nanowires comprise the PND. 
	
	Figure \ref{fig:spiderweb_1} presents a neuron design well-suited to a system with a few tens and possibly hundreds of connections. We refer to this device as the spiderweb neuron. In this design, all upstream signals are combined on a single waveguide. This waveguide enters a spiral region in which it passes a number of SNSPDs which can be wired in series or parallel. Photon wave packets can pass several tens of SNSPDs several tens of times. The system can thus be engineered to spread the absorption probability evenly over the SNSPDs. In Fig. \ref{fig:spikeProbability}, the photons were assumed to pass each nanowire 100 times with a probability of absorption of 1\% at each pass. The size of the detector portion of this neuron can be made as small as $10 \mu$m $\times 10 \mu$m and will depend on the thresholding number of photons. For a threshold of 1,000 photons, the device will be approximately $35 \mu$m $\times 35 \mu$m. The model is discussed in more detail in Appendices \ref{apx:SNSPDAppendix} and \ref{apx:time}, and other neuron designs are discussed in Sec. \ref{sec:dendriticArbor}. In the calculations of Fig. \ref{fig:spikeProbability} it was assumed all photons arrive in a short pulse so nanowire rebiasing dynamics can be neglected. The complex dynamics of the PND receiver array in the case of arbitrary photon arrival times will be the subject of future investigation.
	
	\subsection{\label{sec:series}Differentiable response circuit}
	\begin{figure} 
		\centerline{\includegraphics[width=5.0cm]{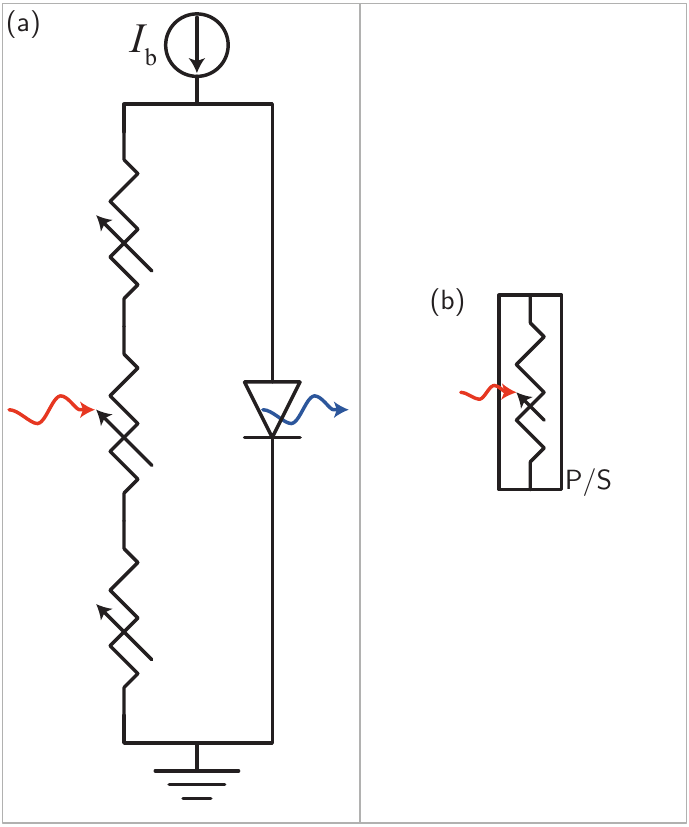}}
		\caption{\label{fig:circuits_SND} (a) SND circuit. (b) Component diagram indicating either SND or PND array. This circuit symbol will be used throughout this article.}
	\end{figure}
	In biological systems, the neuron response is not that of a step function, but rather a nonlinear response taking the form of a sigmoid. For certain neural network back-propagation algorithms, it is important that the response be continuous and differentiable \cite{bi1994}. Figure \ref{fig:circuits_SND}(a) shows the series nanowire detector (SND)  \cite{jafi2012} circuit which achieves a continuous and differentiable nonlinear response. In Fig. \ref{fig:circuits_SND}(b) we define a general optoelectronic circuit element symbolizing either the PND  (Fig. \ref{fig:circuits_PND}) or the SND (Fig. \ref{fig:circuits_SND}). We envision the SND as a single length of superconducting wire with incident photons spread along the length of the wire. As in Fig. \ref{fig:circuits_PND}(a), the detector array is in parallel with the LED. When a single photon is absorbed by the SND, a length of normal wire, called a hotspot, emerges in series with the superconductor, leading to current redistribution between the two branches of the circuit. For common SNSPD materials, this resistance is $\approx 1 \mathrm{k}\Omega$ for typical wire width, while the length of the single hotspot is on the order of $100$ nm \cite{enlo2015,mast2016}. As more photons are absorbed, more hotspots are created and the resistance of the SNSPD increases. This causes the voltage across the LED to increase, and sufficient current can be driven through the diode to produce an optical signal. 
	
	While attempts have been made to utilize this effect for number-resolving single-photon detection \cite{jafi2012}, we emphasize that we propose to utilize this circuit in a very different operating regime. To detect a single photon with near-unity efficiency, an SNSPD is driven close to its critical current, and the ensuing voltage pulse is measured across a 50 $\Omega$ resistor in parallel with the SNSPD. When a photon is absorbed, a 1 k$\Omega$ hotspot is produced, and nearly all current is diverted to the 50 $\Omega$ load. For the application at hand, the device is not intended to observe events of one or a few photons, but rather hundreds to thousands. Thus, diverting the current through a high-impedance diode with $I-V$ approximated by Eq. \ref{eq:pn} enables thresholding with some dynamic range for higher numbers of absorbed photons. The model of this SND-based neuron considers simple joule heating behavior in that each photon absorption event results in the same hotspot resistance, when in reality this will depend on the current through that branch of the circuit, which depends on the temporal dynamics of preceeding absorption events. A thorough study of these dynamics will be the subject of future work.
		
	\begin{figure} 
		\centerline{\includegraphics[width=7.0cm]{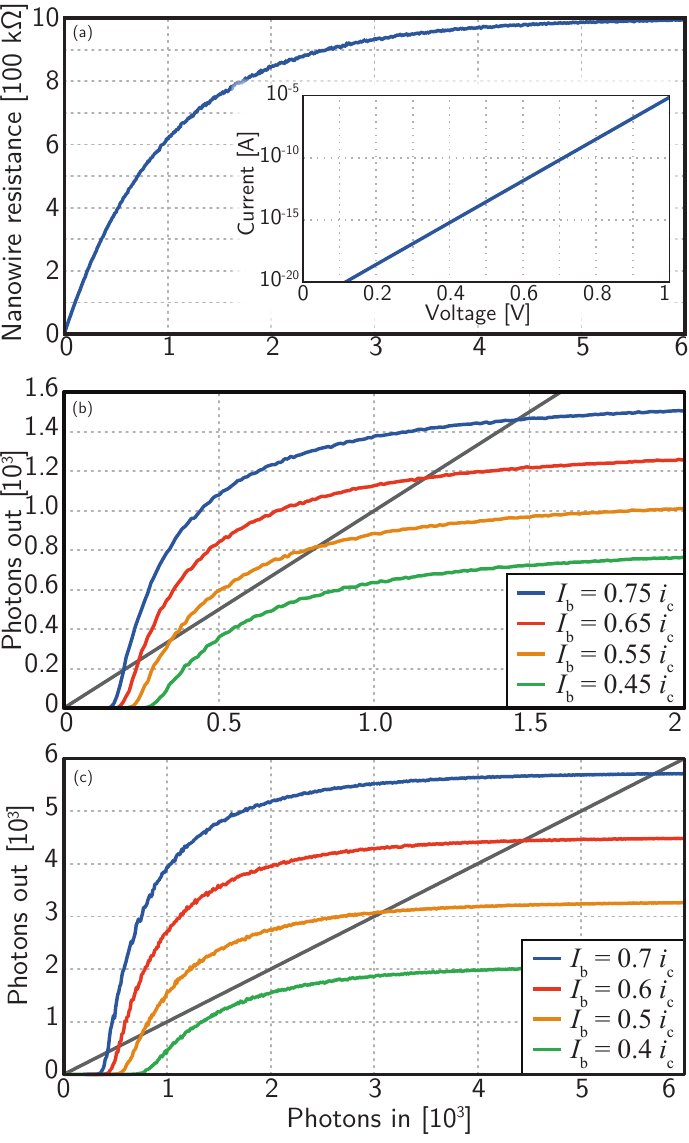}}
		\caption{\label{fig:SNDPlots} Electrical characteristics for SND with $l_{wire}$ = 100 $\mu$m (a) Resistance vs. number of photons for the SND. Inset shows the exponential current-voltage curve for the LED. Photons out versus photons in for SNDs with (b) $i_c$ = 4 $\mu$A, $\eta$ = 1\% and (c) $i_c$ = 8 $\mu$A, $\eta$ = 0.1\%. Here $\eta$ is the efficiency of the LED.}
	\end{figure}	
	Electro-optic performance of the SND is analyzed in Fig. \ref{fig:SNDPlots}. The nanowire resistance as a function of number of absorbed photons is shown in Fig. \ref{fig:SNDPlots}(a). In this model, we assume the photons are incident upon a length of out-and-back nanowire \cite{spga2011,pesc2012,feka2015,namo2015,scgu2016,saga2015,shbu2016} with 100 $\mu$m attenuation length, and it is assumed that two photons absorbed at the same location along the nanowire gives rise to the same resistance as a single photon absorbed at that location. For this reason, the nanowire resistance levels off as a function of number of absorbed photons. The current-voltage relationship of the LED is highly non-linear, as shown in the inset, but above a certain number of absorbed photons the entire length of the absorbing region of the superconductor has been driven normal, and the absorption of additional photons results in no additional resistance, as shown in Fig. \ref{fig:SNDPlots}(b) and (c). Hence, the device has an input-output relationship with an exponential turn-on when a threshold number of photons has been absorbed followed by a flattening of the output when the entire SND has been driven normal. Figures \ref{fig:SNDPlots}(b) and (c) show the photon input-output relationship for two different nanowire designs with critical currents of 4 $\mu$A and 8 $\mu$A respectively, demonstrating the ability to tune the response in hardware. Note that the photon input-output relationship depends on the refractory period, as discussed in Appendix \ref{apx:time}.
	 
	Based on the analysis of Fig. \ref{fig:SNDPlots}, in the SND-based neuron, the normal-state resistance of the SND and the applied bias determine the maximum voltage that can be achieved across the LED. This, in conjunction with the optoelectronic design of the LED, determines the number of photons generated, in contrast to the case of the PND where the number of photons generated is a step response determined by the bias current.
	
	Both the PND-based integrate-and-fire circuit of Fig. \ref{fig:circuits_PND}(a) and the SND-based continuous-response circuit of Fig. \ref{fig:circuits_SND}(a) may offer utility for neuromorphic computing. For the case of the PND, the number of nanowires in the array will be on the order of the number of photons required for threshold. This will also be the order of the number of connections each processing unit makes to other units. Biological systems reveal that scaling to systems with thousands of connections per neuron is desirable \cite{haah2015}. To achieve this number of parallel receiver elements, several geometrical configurations can be utilized to arrange $\approx$1,000 micron-scale SNSPD elements, and the exploration of this design space will be the subject of future work. 
	
	The SND device straightforwardly lends itself to hundreds or thousands of connections. In this case we can expect the thresholding number of photons to be $\approx1,000$, and therefore we would like a nanowire with the length of $1,000$ hotspots. Given the hotspot length of 100nm, the entire length of the nanowire will be on the order of $100 \mu$m, as simulated in Fig. \ref{fig:SNDPlots}. Such a length becomes quite compact when coiled in a spiral [see Fig. \ref{fig:stingray} (b)], and as we will discuss in Sec. \ref{sec:dendriticArbor}, this configuration is well-suited to receive inputs from hundreds to thousands of waveguides.  We will discuss the energy requirements of the SND and PND circuits in Sec. \ref{sec:energy}.

	\subsection{\label{sec:nTron}The nTron current amplifier}
	Introducing an amplifier into the circuits described in Secs. \ref{sec:parallel} and \ref{sec:series} allows decoupling of the firing threshold and LED gain. In a superconducting circuit, this can be done using the nTron, a three terminal supercurrent amplifier \cite{mcbe2014}. When the current in the gate terminal exceeds the critical current, the path from the source to drain is driven normal, diverting the bias current to the parallel load. This recently developed device has been used to drive loads of tens of kilohms, making it suitable for this application.
	
	In Fig. \ref{fig:circuits_variants}(a) we show a variation on the circuit of Fig. \ref{fig:circuits_PND}(a), but instead of driving the same current $I_1$ through the LED after firing, this circuit utilizes an nTron current amplifier to provide gain to the light emitter. This allows us to decouple the current used to bias the receiver from the number of photons produced in the firing event.  Note that in this configuration $I_2$ can be less than $I_1$, making it possible to cover a broad range of input-output responses. The circuit of Fig. \ref{fig:circuits_variants}(a) also expands the state space in which information can be encoded. 
	
	\subsection{\label{sec:cellVariants}Other neuromorphic circuits}			
	We have introduced the basic neuromorphic circuits in secs. \ref{sec:parallel} and \ref{sec:series}. We now introduce several variants on those cells which enable diverse functionality desirable for neuromorphic computing. 
				
	\begin{figure} 
		\centerline{\includegraphics[width=7.0cm]{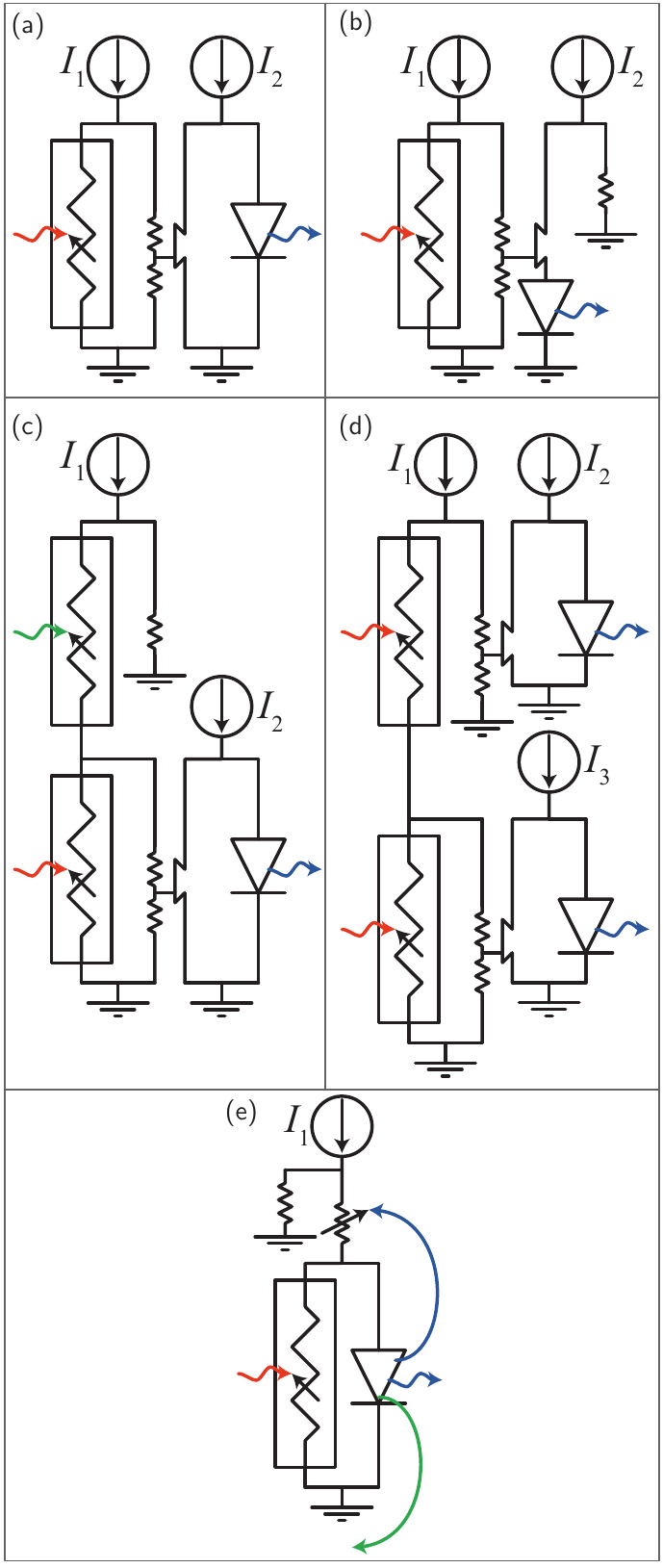}}
		\caption{\label{fig:circuits_variants} Various neuromorphic circuit configurations. (a) PND with nTron amplifier. (b) Integrate and stop firing. (c) Neuron with possibility for both excitatory and inhibitory excitation. In this figure, green corresponds to photons inhibiting firing and red to photons exciting firing. These photons can have different colors. (d) Firing of the upper neuron inhibits firing of the lower neuron. (e) Circuit for achieving self- and upstream-feedback.}
	\end{figure}
	Figure \ref{fig:circuits_variants}(b) shows an alternate configuration in which the LED is being driven by current $I_2$ until a firing event occurs and cuts off the current supply. This circuit is shown with the LED below the nTron, but it could also be implemented without an nTron. Integrate-and-stop-firing neurons such as this can be useful in neuromorphic architectures to provide a means of stimulating various regions of the cortex until a certain level of activity is reached, at which point the firing neuron is quenched.
		
	Another essential functionality of neuromorphic circuits is that of inhibitory connections \cite{sqbe2008,nora2007}. Most neuronal connections provide feedforward excitation wherein an action potential produced by upstream neurons increases the probability of action potentials being produced by downstream neurons. But biological systems also exhibit connections wherein the firing of upstream neurons suppresses the probability of firing events by downstream neurons. Figure \ref{fig:circuits_variants}(c) shows a configuration which achieves this. The lower portion of the circuit is identical to that of Fig. \ref{fig:circuits_variants}(a), but the current $I_{1}$ feeding the receiver first passes through a preliminary nanowire array. Absorption of photons in this region of the circuit reduces the current through the primary receiver, increasing the threshold photon number. Waveguides from different upstream neurons could be routed to these two different ports to establish inhibitory or excitatory connections. In Fig. \ref{fig:circuits_variants}(c) the inputs to the two receivers are drawn with different colors, emphasizing the possibility that integrated photonic filters placed before the neuron could be employed to route different frequencies to the two receivers. With this approach we can employ the use of color to perform inhibitory or excitatory functionality in much the same way that different neurotransmitters perform inhibitory or excitatory functions in biological systems \cite{sqbe2008}. We note that low-loss spectral filters performing this function are commonplace in many integrated-photonic applications.
	
	From an architectural standpoint it may also be useful to establish purely electrical inhibitory connections. In Fig. \ref{fig:circuits_variants}(d) we show a circuit in which two neurons, each with only a single excitatory port, are connected in series. In this configuration, firing events in the upper neuron inhibit firing events in the lower neuron. Such a configuration is useful for moderating the net firing activity of groups of neurons.
			
	It is also advantageous to have a means by which a single neuron can moderate its own firing activity. Such behavior is straightforward to implement, as is shown in Fig. \ref{fig:circuits_variants}(e). A power tap is added to the output of the LED, and some fraction of the produced light is incident upon a receiver in series with the current supply to the receiver array. The superconducting wire in this location may be wider than the integrating receiver, and it therefore may be designed to quench the current only when a large number of photons drives the superconducting wire normal. 
		
	In addition to self-feedback, biological neurons send both downstream signals as well as upstream signals when an action potential fires. The upstream signals are believed to be critical for spike-timing-dependent plasticity (STDP) and synchronization of circuit behavior via threshold modification. To briefly hint at how this may be implemented in the proposed platform, the green arrow leaving the LED in Fig. \ref{fig:circuits_variants}(e) indicates that a power tap could also be used for upstream feedback. The color of this arrow is meant to remind us that it may be advantageous to use different frequencies of light for downstream and upstream signaling. An LED could be fabricated to emit at two distinct wavelengths or across some region of bandwidth, and integrated spectral filters could be employed to route the two signals. Alternatively, two different LEDs coupled to two different waveguides could be utilized.
		
	In this section we have presented several superconducting optoelectronic neuromorphic circuits covering a wide range of functions. We refer to members of this class of circuits as single-photon optoelectronic neurons (SPONs). We now proceed to discuss additional aspects of their performance.

	\subsection{\label{sec:energy}Energy consumption}
	\begin{figure} 
		\centerline{\includegraphics[width=7.0cm]{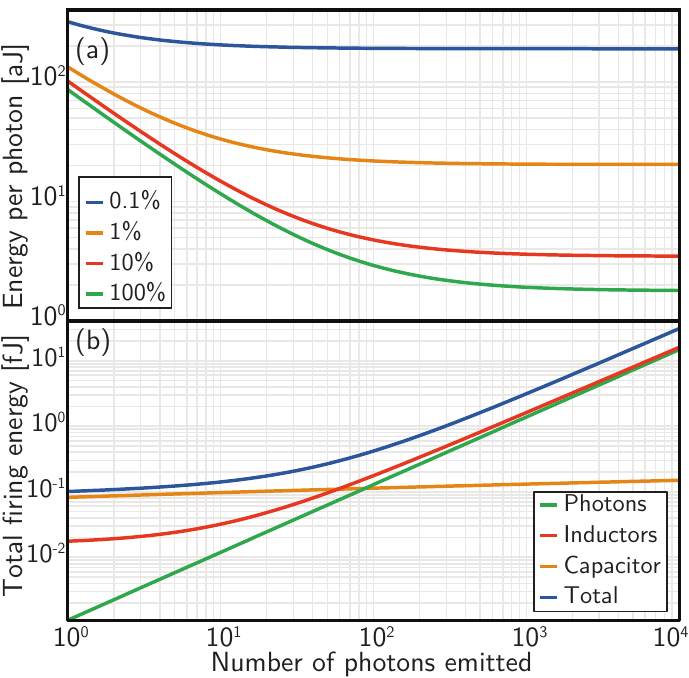}}
		\caption{\label{fig:ledPlots}(a) Energy required to generate a single photon versus number of photons emitted for four different LED efficiencies. (b) Contributions to total energy consumption for a 10\% efficient LED.}
	\end{figure}
	We have introduced the basic SPON circuits of the proposed neuromorphic computing platform, and we are in a position to estimate the energy required for a firing event. A complete neuron firing event involves supplying current to the inductors associated with all superconducting wires (including the detectors), charging the capacitor associated with the LED $p-i-n$ junction, and driving current through the LED to produce light. For the case of the PND circuit of Fig. \ref{fig:circuits_variants}(a), we analyze the energy consumption of each of these three contributions. 
	
	In this model, we assume one inductor $L_{\mathrm{SNSPD}}$ in the PND array for each photon, as well as a series inductance to achieve the desired temporal response (see Appendix \ref{apx:time}). We assume each element of the PND is 500 squares, while the entire receiver array is in series with 5,000 squares of inductance. At low photon numbers the energy consumption from inductance is dominated by the series inductance, but for higher numbers it is dominated by the PND array and grows linearly. The energy required for photon production is calculated simply as $E_g n_{\nu}/\eta$, where $E_g$ is the band gap of Si, $n_{\nu}$ is the number of photons created, and $\eta$ is the efficiency. Thus, within this model, the contribution to energy consumption due to photon creation is linear throughout. We use $E_g$ in this model to because it is an upper bound on the photon energy. Any photon transmitted through a Si waveguide will have energy below the band gap. We assume a superconducting material with a sheet inductance of 400 pH/$\square$ (such as WSi), and a parallel-plate capacitive model for the LED as described in Appendix \ref{apx:LEDAppendix}. 
	
	In Fig. \ref{fig:ledPlots}(a) we plot the total energy per photon as a function of the number of photons emitted for four values of LED efficiency. We find that with a unity-efficiency LED, the energy per photon could be as low as 2 aJ when larger photon numbers are created. This remarkably low number is still an order of magnitude greater than the 0.16 aJ stored in the $h\nu$ of the light quantum itself (assuming $\lambda = 1.22 \mu m$), with the extra energy going to supplying current to the inductors and charge to the capacitor. The figure reveals that producing LEDs with efficiency above 10\% has only a modest benefit, as the contribution to energy consumption from inductance will become the limiting factor. However, for thresholding on larger photon numbers, as would be desirable for neurons with more connections, the inductance per photon can likely be reduced. While a 100\% efficient LED may not be realized, even a 1\% efficient LED leads to 20 aJ/photon. This energy efficiency illustrates the promise of superconducting electronics and faint-light signals.
	
	In Fig. \ref{fig:ledPlots}(b) we show the contribution to the total energy from the various circuit elements for the case of a 10\% efficient LED. This efficiency is chosen for this plot because it is the value at which the contributions from inductance and photon production are nearly equal for photon numbers near or above 100. For low photon numbers, the dominant contribution is in charging the LED capacitor. Due to the highly nonlinear LED current-voltage relationship, a small increase in the voltage across the LED leads to a large gain in current. The capacitive energy is nearly constant across the range of photon numbers considered here, and for larger photon numbers it makes a negligible contribution. 
	
	In the case of the case of the SND circuit of Fig. \ref{fig:circuits_SND} with parameters as shown in Fig. \ref{fig:SNDPlots} (b) driven at 0.6$I_c$ and receiving $10^3$ photons, and assuming a hotspot recovery time of 50 ns and an LED with 1\% efficiency, the device achieves 100 aJ/synapse event. While not as efficient as the PND neuron, this device design still lends itself to massive scaling, as will be discussed in Sec. \ref{sec:scaling}. 
	
	We believe an LED with 1\% system efficiency is realistic in a nanophotonic environment at cryogenic temperature and with faint light levels desired. Therefore, we use 20 aJ/photon as a representative number for what this platform can hope to achieve. We use the energy per photon as the energy per firing event per synapse (commonly referred to as the energy per synapse event), because the goal of the system is to produce neurons which threshold on a number of photons roughly equivalent to the number of connections made by the neuron. A neuron receiving 100 signals from upstream will threshold on 100 photons. It will produce 100 photons in a firing event, and distribute them amongst 100 downstream synapses. Therefore, the energy per synapse event is calculated as the total energy of the firing event divided by the number of connections. In our case, for systems with 100 to 10,000 connections per unit, 20 aJ/synapse event is a realistic number. 
	
	The second law of thermodynamics informs us that to keep a system at 2 K, the cooling power required is 150 W/W. Assuming a $15\%$ efficient cooling system, this gives an estimate of 1 kW/W. Multiplying our conservative estimate of 20 aJ/synapse event by this factor of $10^3$, the hardware achieves an energy consumption of 20 fJ/synapse event, a value which is competitive with or better than any hardware demonstrated to date. Similarly, while the human brain uses 20 W to perform roughly $10^{14}$ synapse events per second, a power budget of 20 W, corresponding to 20 mW of device power, would enable our system to achieve $10^{15}$ synapse events per second. Success in developing LEDs with higher efficiency, reduction of the device inductance, and utilization of superconducting materials operating at higher temperatures would further increase the advantage. Additionally, while transistor technologies inevitably leak current, superconducting devices can be engineered to draw no power in the steady state and can be DC biased without loss using Josephson junctions \cite{kisa2011}.

	\subsection{\label{sec:LED}Electrically-injected light source}
	Having introduced the proposed optoelectronic neuronal circuits, we now proceed to analyze the operation and performance requirements of the LED. As discussed in the previous section, we target operation efficiencies of around $10\%$. This efficiency is relatively easy to attain in III-V semiconductors such as GaAs and InP. However, for the application at hand, massive scaling is a priority, and this requires photonic-electronic process integration. A single source with 100\% efficiency is less desirable than the ability to scale to millions (and eventually billions) of sources each with 1\% efficiency. We also require low loss waveguides with the potential for reconfigurability (see Sec. \ref{sec:connectivity}).
	
	One option is to implement these devices on a GaAs or InP substrate. These have been the materials of choice for photonic integrated circuits where light sources are of the utmost importance. Quantum dot/well LEDs/lasers can be electrically injected with high efficiency on this platform \cite{Nagarajan2010} and combined with high index (III-V) waveguides to form the synaptic connections described in Sec. \ref{sec:dendriticArbor}. Another option would be to implement the light sources in the III-V material, and then couple to low-temperature deposited materials with low-loss waveguides \cite{Biberman2011,shbu2016} such as a-Si or SiN. A III-V platform has the advantage of high efficiency light sources, but massive scaling on III-V substrates has historically been more difficult and expensive than on Si substrates. This drawback, while not fundamental, may prove significant in halting the development of this technology, especially since high emitter efficiencies are not a strict requirement for neuromorphic computing.
	
	Another option is hybrid III-V/silicon integration. Hybrid III-V/silicon has followed one of three approaches \cite{Zhou2015}: direct mounting, wafer bonding, or III-V material grown on Si. While direct-mounting or wafer bonding are currently the preferred methods for optical interconnect applications, these applications typically require a single source that can be diverted to multiple components. For the proposed neuromorphic computing platform, we desire a separate electrically-injected source for each neuron. Direct mounting therefore is not an option, but wafer bonding may be able to achieve the yield and reproducibility required for this application. Direct hetero-epitaxial growth offers the most promise for hybrid integration with this system. In this case, the desired light source would be templated III-V quantum dots grown in the intrinsic region of a lateral Si $p-i-n$ junction. While great progress in this field has been made \cite{yuka2002,beat2006,sh2007,mima2012,lizh2014,wali2015,scbo2015}, additional effort is needed to achieve the waveguide-integrated sources required for this system. Promisingly, electrically injected single-photon emission has been demonstrated in these materials \cite{yuka2002,beat2006,sh2007,mima2012}. While single-photon emission is not a requirement for the present application, a desirable property of the emitters is that they have low photon number variance (defined as the standard deviation of the number of photons output for a given input current pulse over an ensemble of measurements). The fact that single-photon emission has been demonstrated in various systems indicates the possibility to bring this photon number variance down to the range of a few photons. 
	
	A major disadvantage of this hetero-epitaxy approach is the significant cost and difficulty associated with growing these materials. As this approach matures, this material platform may become more desirable. Another similar approach using Ge \cite{Liu2010b} or Ge quantum dots \cite{Zeng2015} may also prove useful.
	
	A commonly overlooked light source that may prove particularly promising for this application is emissive centers in Si \cite{da1989}. These have proved unattractive for optical interconnects due to very low efficiencies at room temperature. Much work in this area was motivated by the prospect of room-temperature light sources \cite{reca2009} for CMOS and telecommunications \cite{suwa2015}, and in particular room temperature lasers. This includes various point defects in Si including Er \cite{enpo1985,paga1996,hase2001,pusu2013} and other emissive centers giving rise to electric-dipole-mediated transitions \cite{pa1978,dali1987,brbr1989,da1989,clko2005,rosh2007,rosh20072,bata2007,yaba2010,suku2014}, as well as band-edge or Si nanocrystal-based emission processes \cite{nglo2001,grzh2001,wabo2005}. While the efficiencies of many of these emitters fall off exponentially with increasing temperature, the SNSPDs required for this application operate at cryogenic temperatures where many point defects have suitable efficiencies. A large number of emissive centers are under consideration for this application \cite{da1989}. 
	
	The main challenge is the successful integration of large numbers of emitters with the ultimate goal being billions integrated in a system. Many emissive centers can be easily fabricated in a CMOS-compatible process via ion implantation and annealing \cite{dali1987,da1989,cogo1999,shxu2007,bata2007,yaba2010,suku2014}.
	\begin{figure} 
		\centerline{\includegraphics[width=7.0cm]{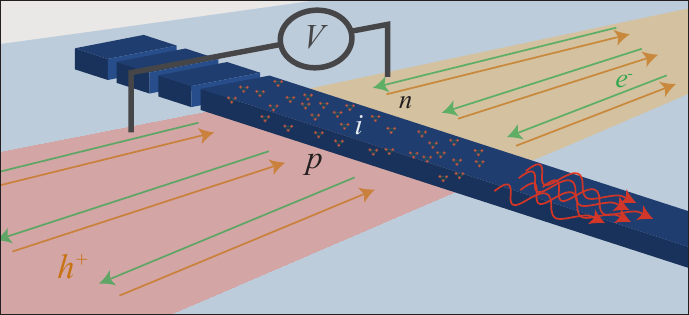}}
		\caption{\label{fig:ledSchematic} Schematic of a monolithically integrated electrically-injected emissive center LED in Si for the proposed neuromorphic computing application.}
	\end{figure}
	A schematic of the desired device is depicted in Fig. \ref{fig:ledSchematic}. A $p-i-n$ junction is created in a ridge waveguide. Emitters are located only in the ridge (intrinsic) region via lithographic patterning, and light is obtained from forward biasing the junction. While this is a relatively standard configuration of an LED, for the application at hand it is important to keep the emitters localized only in the intrinsic region of the LED as their presence elsewhere in the waveguides would lead to intolerable loss. Thus, the ability to lithographically control the location of emitters is crucial.
	
	With co-implantation of multiple impurities it is possible to add additional (color) degrees of freedom to the platform. Similarly, on a III-V platform we could take advantage of inhomogeneous broadening of the quantum dot spectrum and tuning of dot size via templating or growth conditions.
	
	We note that the neuromorphic computing platform proposed here is not tied to any one of these light sources, and indeed there are other possible light sources that we have not discussed. For the calculations throughout the present work, we have assumed LEDs with 1\% efficiency at 1.22 $\mu$m in a waveguiding medium with index of 3.52 with a cladding of 1.46 above and below.
					
	\subsection{\label{section:summary}Summary}	
	We have now presented several superconducting optoelectronic circuits capturing a broad range of neuromorphic behaviors. We have presented basic thresholding SPON circuits of Figs. \ref{fig:circuits_PND} and \ref{fig:circuits_SND}; variants on these circuits as shown in Fig. \ref{fig:circuits_variants} which enable gain, integrate-and-stop, and inhibitory connections; and circuits with self- and upstream-feedback, as shown in Fig. \ref{fig:circuits_variants}(e). We now discuss the means by which we propose to connect these processing units. 
			
	\section{\label{sec:connectivity}Connectivity}
	Of central importance to the implementation of the proposed neuromorphic platform is the network of waveguides that connect the processing units. Optical waveguides offer the possibility for improved performance over electrical connections by allowing individual neurons to integrate signals from many sources without the need for time-multiplexing. Due to the additional energy cost associated with the capacitance of additional wires \cite{Goodman1985}, electrical neurons must utilize shared wires. Voltage pulses from different neurons on the same bus will interact. To prevent this, pulses must be delayed in time. 
	
	In the following section we will discuss how a network of optical waveguides can be implemented to form the connections between the SPON circuits presented in Sec. \ref{sec:SPON}. Each neuron will have a waveguide exiting the LED and leading to many branching waveguides, which we liken to the axon and its arbor, and another set of integrating waveguides combining signals received from upstream neurons, which we liken to the dendritic arbor, as shown schematically in Fig. \ref{fig:cartoon}. Connections between these input and output waveguides act as synapses in this network. We outline a mechanism for varying the strength of the connections between various input and output waveguides, which is similar to varying synaptic weights in biological systems. We emphasize that other methods of connecting neurons in three dimensions using the same optoelectronic neurons are also possible. One can envision using gratings, flat lenses \cite{Fattal2010}, metasurfaces \cite{Kildishev2013,Yu2014}, or optical phased arrays \cite{Doylend2011,Sun2013} to direct signals between neurons. Additionally, electrical means of changing synaptic weights at the receivers may prove useful.
	 
	\subsection{\label{sec:dendriticArbor}The dendritic arbor}
	The dendritic arbor of a neuron collects signals from upstream neurons. For optoelectronic neurons, the equivalent of this is a waveguide network that combines optical signals from many other neurons to the neuron for detection. At each neuron, the device must be designed to combine the modes from a large number of waveguides on a PND or SND with low loss. There are likely many ways to achieve this functionality, and here we explore two. 
	 												
	\begin{figure} 
		\centerline{\includegraphics[width=7.0cm]{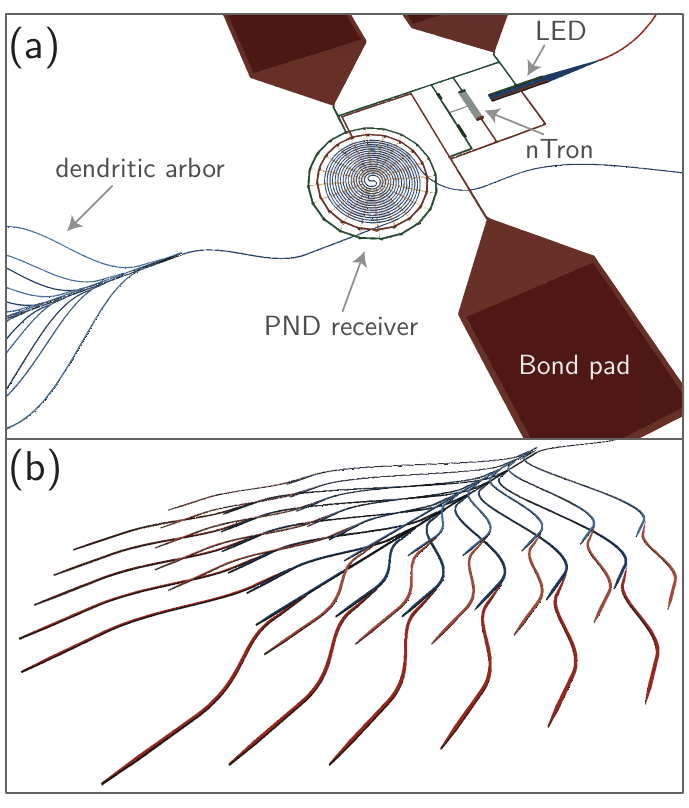}}
		\caption{\label{fig:spiderweb_2} The spiderweb neuron. (a) Overview of device. (b) Dendritic arbor design which combines light from multiple neurons.}
	\end{figure}	
	A schematic of the first approach is presented in Fig. \ref{fig:spiderweb_2}(a), showing the spiral waveguide receiver of the spiderweb SPON, the nTron, and the LED emitter. The major challenge of this device design is the merging of many single-mode waveguides into one multi-mode waveguide which enters the spiral. The proposed technique for accomplishing this is shown in Fig. \ref{fig:spiderweb_2}(b). Two single-mode waveguides cannot be combined into one single-mode waveguide without significant loss \cite{sthe2015}. However, two single mode waveguides can be combined into one dual-mode waveguide nearly losslessly. In Fig. \ref{fig:spiderweb_2}(b), several single-mode waveguides combine their power on a given main spine. That spine can receive at its input one single mode. As it continues to receive more modes, its width must grow. The lower-order modes of this adiabatically tapering multimode waveguide can pass each new single-mode input nearly losslessly as long as the width of the spine has grown to support an additional mode by the location of the next input waveguide. More detail regarding the optical design of this structure is given in Appendix \ref{apx:wgAppendix}. Modal simulations reveal that a waveguide width of 2 $\mu$m in 200 nm-thick Si is sufficient to support several tens of modes at 1220 nm wavelength, each with tolerably small bending loss with a 10 $\mu$m radius of curvature. Therefore, this dendritic arbor and receiver design is suitable for the compact combining signals from $\approx 40$ upstream neurons. 
				
	\begin{figure} 
		\centerline{\includegraphics[width=7.0cm]{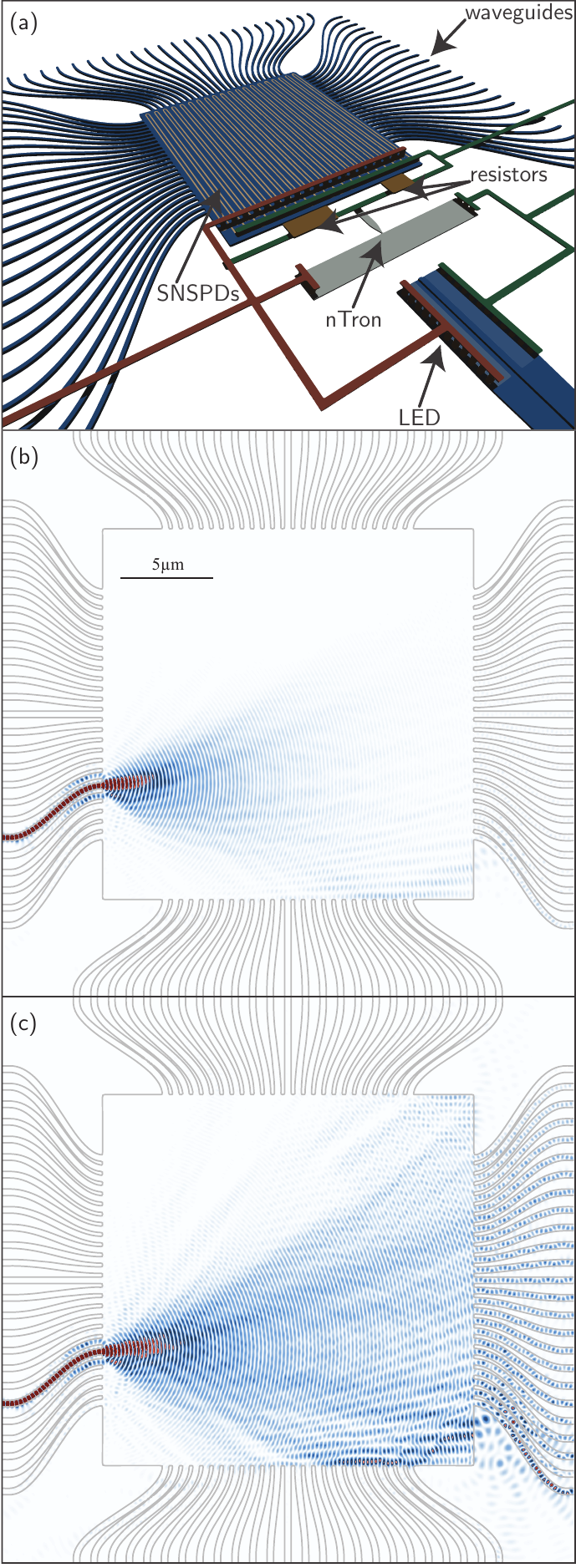}}
		\caption{\label{fig:stingray} (a) Schematic overview of the stingray neuron. (b) FDTD simulation of the dendritic arbor for the stingray neuron with SNSPDs present to absorb the light and (c) without SNSPDs present.}
	\end{figure}
	The second proposed design is better suited to scaling to larger numbers of inputs. It is shown in Fig. \ref{fig:stingray}. In this design, referred to as the stingray SPON, the input waveguides are directly combined on a landing pad housing the PND or SND array. The implementation with a PND is shown in Fig. \ref{fig:stingray}(a). As is shown in Appendix \ref{apx:wgAppendix}, the minimum spacing required to avoid modal coupling is 600 nm at the input of the cell. From these input ports, the waveguides enter an array of sine bends where their spacing is reduced to enter the smaller landing pad containing the nanowires. In this sine region, inter-modal coupling is tolerated (and perhaps even desirable to spread the photons across the nanowires), as all waveguides ultimately terminate on the detector array. Figure \ref{fig:stingray}(b) and (c) shows 2D FDTD simulations of the structure. Figure \ref{fig:stingray}(b) shows the propagation of light into the receiver body in the presence of absorbing nanowires, while Fig. \ref{fig:stingray}(c) shows propagation without the absorbing nanowires. Here, 100 waveguides terminate on a receiver body with less than 0.2 dB insertion loss from any port, with the outer-most ports giving the most loss, and the inner-most ports achieving near zero insertion loss. In this context, insertion loss refers to light entering and leaving the simulation without being absorbed in the nanowire array. Calculated quantitatively with pulsed excitation, we find the majority of loss is due to light scattering and not entering the detector array rather than being transmitted through the receiver due to inadequate absorption. The entire receiver of Fig. \ref{fig:stingray}(b) occupies 30 $\mu$m $\times$ 30 $\mu$m. A design with 204 input waveguides and less than 1 dB insertion loss with a footprint of 60 $\mu$m $\times$ 60 $\mu$m has also been found. For larger numbers of inputs, the simulations become cumbersome. Yet scaling to larger systems is clearly possible. 
		
	For threshold-based computation, processing units with large numbers of connections are advantageous \cite{an1988,haah2015}. Biological systems achieve massive interconnectivity with 3D branching networks and dedicated wires for each connection. To achieve this level of massive interconnectivity, we propose the use of multi-layer photonics. Recent work has demonstrated the utility of low-temperature-deposited dielectrics \cite{shch2016,shbu2016} and superconductors \cite{shbu2016} for scalable integrated photonics. For future massive scaling we propose the use of waveguide routing networks and dendritic arbors spanning several\textemdash and possibly up to tens\textemdash of photonic and superconducting layers. A hybrid of the aforementioned spiderweb and stingray neuron designs could be implemented in which higher vertical mode orders are utilized as well as higher lateral mode orders, and massively multimode waveguides deliver their photon pulses to SNSPD receivers. These receivers could be implemented between waveguiding layers. At present, the technical challenge of building networks with processing units supporting tens to hundreds of connections is a serious one, so we mention the fully 3D, multilayer photonic approach to emphasize that this neuromorphic platform holds promise for scaling far into the technological future, but such sophisticated processing is not required to implement even very advanced systems with 2D interconnectivity supporting hundreds of high-bandwidth connections per unit.

	\subsection{\label{sec:axon} The axon and its arborization}
	The output waveguide (axon) from a unit's LED must split into as many branches as there are connections to be made. While such a power splitter may seem to be the time-reversed case of the dendritic arbor, the initial conditions make this device significantly easier to implement. In the case of the dendritic arbor, one cannot assume the optical field will populate the arbor modes in a particular manner, so while a power splitter can readily couple from a single-mode waveguide into many other single-mode waveguides, multiple single-mode waveguides cannot simply merge their power into a single-mode waveguide unless a particular distribution of power is present in the input waveguides. Such power splitters \cite{lixu2013} can be made with a small footprint and low loss. It is straightforward to generalize such power splitters into the third dimension with multilayer photonics, and such an implementation would enable thousands of synapses with a volume of 10 $\mu$m$^3$/synapse. 
			
	\subsection{\label{sec:plasticity}Learning, reconfiguration, and plasticity}
		
	\begin{figure} 
		\centerline{\includegraphics[width=7.0cm]{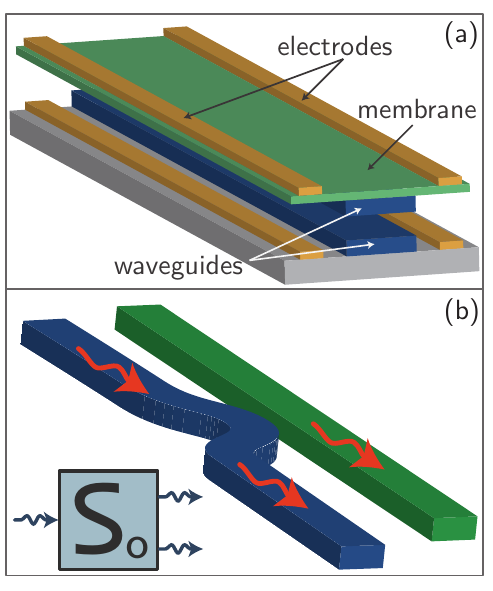}}
		\caption{\label{fig:mechanicalSynapse} Interlayer waveguide coupler with electro-mechanically tunable coupling. The inset shows an abstract representation of the synaptic circuit element that we will use in subsequent network diagrams.}
	\end{figure}		
	An important aspect of any neuromorphic computing system is the ability to establish the strength of interaction between connected units. These connection strengths, often referred to as the weight matrix, are important for memory and learning. This weight matrix determines how much of the light from the firing of a particular neuron is coupled into any other neuron, analogous to the synaptic strength between two neurons in a biological system.
	
	As a first implementation, fixed connection weights are quite useful for many computing applications \cite{mear2014}. This can be readily accomplished by branching the output waveguide from one neuron and routing those waveguide branches to various downstream target neuron input waveguides.
	
	However, while fixed interaction weights are useful as a preliminary tool, one would like to develop a system in which the interaction strengths are variable. This is challenging at cryogenic temperatures, where modulators that rely on either the thermo-optic effect or free carrier injection are ineffective, while electro-optic switches require too much space for this application. We propose the employment of electro-mechanically actuated waveguide couplers, schematically depicted in Fig. \ref{fig:mechanicalSynapse} (a) and (b). The amount of light coupled from one waveguide to the other is determined by the distance between them. These waveguides can be coupled either vertically [Fig. \ref{fig:mechanicalSynapse} (a)] or laterally [Fig. \ref{fig:mechanicalSynapse} (b)]. This distance can be controlled  electro-mechanically, and anywhere from 0\% to 100\% of the light can be coupled from one waveguide to the other. The minimum coupling would be set in hardware, as the gap at 0 V is the maximum. Any applied voltage (positive or negative) produces an attractive force between the two waveguides. We would then like activity within the circuits to build up voltage between the waveguides, and increase the strength of the synapse. Such couplers have recently been demonstrated \cite{sequ2016} in a highly scaled configuration. In Ref. \cite{sequ2016}, 4096 such switches were operated with $> 60$ dB extinction ratio and actuation voltage of 40 V. Due to the relaxed visibility requirements for this application, we expect much lower voltages will suffice. 
	
	To assess the utility of such synapses for neuromorphic computing, one must further specify the target application. To this end, we separate potential applications into two classes, which we will refer to as supervised and unsupervised systems. For supervised systems, an input stimulus is injected into the system, the output is recorded, and the weight matrix is updated through a training algorithm to improve the output relative to a target. For such an application, one anticipates using control electronics to interface with the neuromorphic system, and arbitrary voltages can be applied to the various synaptic elements. 
	
	For more highly scaled implementations emulating the behavior of biological organisms, we turn our attention to unsupervised systems. Here it is important that each synapse be as small as possible to enable massive scaling, but it is also important that voltages be modest, as we would like activity in the circuits to be capable of reconfiguring the synapses. In particular, we would like firing events from upstream neurons followed closely by firing events by downstream neurons to place charge on this MEMS capacitor (waveguide coupler), and thereby decrease the gap between the two waveguides and increase the optical coupling and therefore the synaptic strength. This coordinated charging of the membrane would accomplish spike-timing-dependent plasticity, an important learning and memory reinforcement mechanism in biological neural systems. In this mode of operation, we envision eliminating external control circuits and achieving the capacitor charging using integrated superconducting circuits to distribute current based on photon absorption events. The storage of charge on a capacitor required for this device operation is very similar to dynamic random access memory (DRAM), which is a mature technology. While implementing what is essentially spike-timing-dependent DRAM with suspended waveguide membranes presents a technical challenge, it offers a promising means to implement truly neuromorphic learning within this optoelectronic platform.
	
	While the size of mechanical waveguide couplers and the voltages required for their operation are commensurate with the requirements for scaling this technology, an implementation of variable synaptic weights which does not rely on mechanically mobile components would be advantageous. It may be possible to implement synapses in the electronic domain by making use of superconducting circuit elements or magnetic elements such as magnetic tunnel junctions or magnetic Josephson junctions \cite{vevi2013}. Such an approach to memory will be investigated in future work. Additionally we note that a variable weight could be achieved with a tunable Mach-Zehnder interferometer. However, the size of such devices makes them poorly suited to highly scaled systems.
			
	\section{\label{sec:apps}Networks and scaling}
	We have now discussed neural circuits based on optical signaling. We have discussed various means to connect these optical and electrical signals in a time-varying manner with event-based plasticity. 	
	\begin{figure} 
		\centerline{\includegraphics[width=7.0cm]{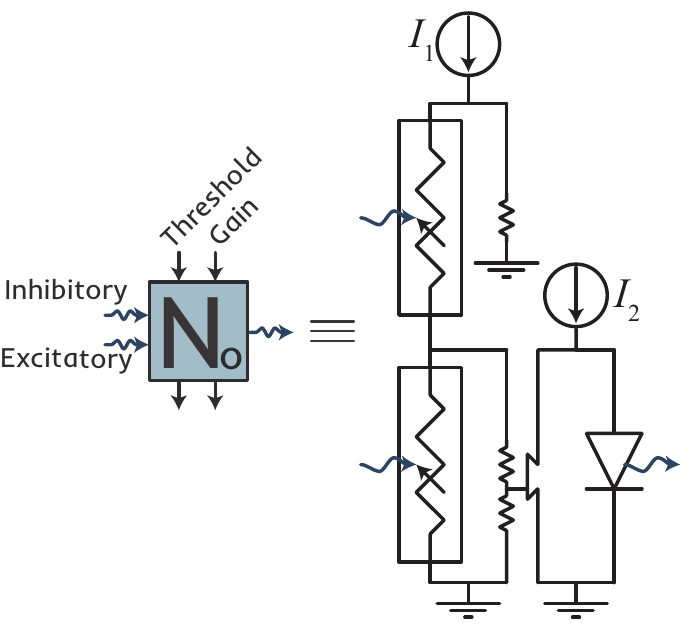}}
		\caption{\label{fig:N}Abstract symbol definition for general neuron with inhibition and gain.}
	\end{figure}
	In Fig. \ref{fig:N} we again show the inhibitory SPON of Sec. \ref{sec:series} and introduce an abstract symbol to represent the circuit, labeled $\mathrm{N_o}$, which will be used in the following sections as an element in networks. We refer to networks comprising interconnected SPONs as superconducting optoelectronic networks (SOENs). In this and the following schematics we represent electrical inputs and outputs as black arrows running vertically and optical inputs and outputs as colored, wavy arrows running horizontally. In Fig. \ref{fig:N}, we emphasize that the optical processing unit can receive and transmit electrical and optical signals each in two ports. The electrical signals affect SPON threshold and gain, while the optical ports are either excitatory or inhibitory. This full functionality need not be employed, and as few as one optical input and output and one electrical input can be utilized.
		
	We will now illustrate how the circuits presented in Sec. \ref{sec:SPON} may be put to use in systems by considering the canonical example of the multi-layer perceptron (MLP) in Sec. \ref{sec:cnn}. This will lead us into a more general discussion of SOEN scaling in Sec. \ref{sec:scaling}. 
				
	\subsection{\label{sec:cnn}Multi-layer perceptron}
	Perhaps the most studied implementation of NNs is the MLP \cite{bi1994}, and its contemporary counterpart, the convolutional neural network (CNN) \cite{sc2014}. Our consideration of the MLP will provide insight into other applications of this platform in terms of important quantities such as speed, size, and dynamic range.
	
\begin{figure} 
		\centerline{\includegraphics[width=7.0cm]{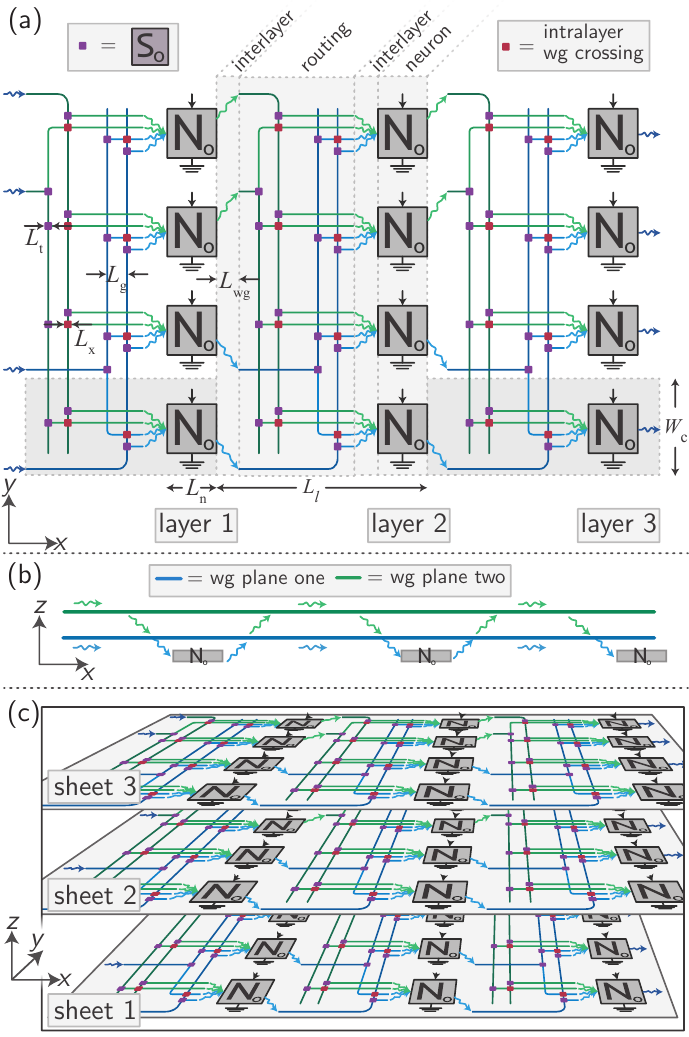}}
		\caption{\label{fig:app_cnn_1} (a) Schematic of the MLP implemented with the SOEN platform. (b) Cross section in the $x-z$ plane. (c) Three-dimensional schematic of stacked die. Part (a) illustrates layers of neurons in the network, part (b) illustrates planes of routing waveguides, and part (c) illustrates sheets of stacked die. }
	\end{figure}

	Generally speaking, the MLP consists of a number of inputs incident on a weight matrix (array of synapses) which feeds into a layer of neurons. The output of this layer of neurons projects to at least one more layer of weights and neurons, and often several, before being output from the system. In Fig. \ref{fig:app_cnn_1}(a) we show a schematic diagram of how such an MLP is likely to be implemented. Such an MLP could be achieved with a single plane of routing waveguides or many such planes. Here we use ``plane'' to refer to vertically stacked dielectric layers to avoid confusion with the processing layers of the MLP, progressing horizontally in Fig. \ref{fig:app_cnn_1}(a). The processing layers of the MLP are labeled in Fig. \ref{fig:app_cnn_1}(a), and the cross sectional view of planes of routing waveguides is shown in Fig. \ref{fig:app_cnn_1}(b). Stacked sheets of die are illustrated in Fig. \ref{fig:app_cnn_1}(c).
	
	Several factors determine the functionality of an MLP. These include the dynamic range of the inputs, the speed with which the inputs can be received, the bit depth of the synaptic weights, and the speed with which the weights can be reconfigured. From Fig. \ref{fig:SNDPlots}(c) we see that for 0.7$I_c$ the response turns on at around 500 photons, and it has roughly leveled out by 3,000 photons. For this case, the dynamic range of the inputs is therefore $\mathrm{log}_2(2500) \approx 11$ bits. The speed with which inputs can be received is limited by the device reset time of 50 ns, so a 20 MHz input rate is achievable. The bit depth of the weights depends on the number of discrete values of coupling achievable between the two waveguides involved in a synapse, and further investigation will be required to report a valid estimate for this number. The speed with which the weights can be changed is at least 1 MHz \cite{sequ2016}. 	 		
	
\begin{figure} 
		\centerline{\includegraphics[width=7.0cm]{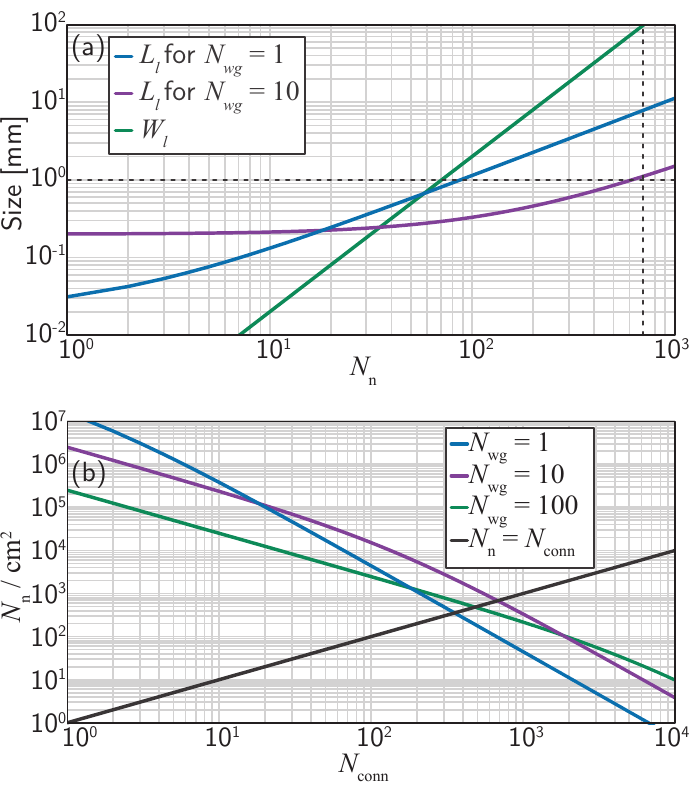}}
		\caption{\label{fig:app_cnn_2} (a) Length and width of layer versus number of neurons in a layer assuming each neuron in a given layer is connected to each neuron in the next layer. (b) Number of neurons per cm$^2$ versus number of connections per neuron.}
	\end{figure}

	The number of inputs, the number of connections per neuron, and the number of MLP layers all affect the size and complexity of MLP that can be fabricated on a given die. In Fig. \ref{fig:app_cnn_2}(a) and (b) we consider a model of these factors to estimate what may be achieved with reasonable size. Figure \ref{fig:app_cnn_2}(a) assesses the length, $L_l$, and width, $W_l$, of a single MLP layer, as given by Eq. \ref{eq:length} as a function of the number of neurons in an MLP layer, $N_n$, for two different values of the number of vertically stacked waveguide planes, $N_{\mathrm{wg}}$. The model assumes a feedforward configuration wherein every neuron in a given MLP layer is connected to every neuron in the next MLP layer with a variable-weight connection.  The total width of an MLP layer is also plotted. See Appendix \ref{apx:scaling} for more information. If we assume that a 10 cm $\times$ 10 cm die is the largest we would like to fabricate, we find the width limits the number of connections per neurons to 700, and we are thus considering MLP layers with 700 inputs and 700 neurons per layer. For the case with $N_{\mathrm{wg}} = 10$, the length of an MLP layer with 700 connections per neuron is 1 mm. We can therefore fit 100 such MLP layers on the 10 cm $\times$ 10 cm die. The total number of neurons would be 70,000. An MLP or CNN with 700 inputs, 700 connections per neuron, and 100 layers receiving inputs at 20 MHz with weight reconfiguration speed of 1 MHz would be a very powerful tool. While it is not necessarily optimal to work with a neural network of 100 layers as shallower networks are advantageous for several reasons \cite{an1988}, we present this model to quantify SOEN spatial scaling keeping in mind that network depth can be traded for a larger number of inputs or larger connectivity. As a point of comparison, the recent demonstration of a computer defeating the world champion Go player input the state of the board as a $19\times19$ matrix (361 inputs) to the 13-layer deep neural network \cite{sihu2016}. The bit depth of the synapses proposed here is unlikely to reach the 32 bits utilized in software implementations running on modern GPUs, but there are likely many applications in which such a constraint is minor compared to the system advantages of speed, complexity, and connectivity. 		
	
	\subsection{\label{sec:scaling}Scaling}
	To further pursue this discussion of the scaling of the MLP (or other similar neuromorphic computing systems), we consider the number of neurons in an area of one cm$^2$ versus the number of connections per neuron, $N_{\mathrm{conn}}$. Figure \ref{fig:app_cnn_2}(b) shows the results of the model of Eq. \ref{eq:length} for $N_{\mathrm{wg}} = 1, 10,$ and $100$. If $N_{\mathrm{conn}}$ = 10 is sufficient for a given application, we can achieve a neuron density of 400,000 neurons per cm$^2$. Due to the size of interlayer couplers, this is achieved more compactly with $N_{\mathrm{wg}} = 1$ than with $N_{\mathrm{wg}} = 10$. For $N_{\mathrm{conn}}$ in the range of 100 to 1,000, it becomes advantageous to utilize $N_{\mathrm{wg}} = 10$. For $N_{\mathrm{conn}} = 100$, over 10,000 neurons will fit within a cm$^2$, and for $N_{\mathrm{conn}} = 1,000$, 300 neurons will fit within a cm$^2$. It does not become advantageous to use $N_{\mathrm{wg}} = 100$ until $N_{\mathrm{conn}} = 2,000$, and even then the gain is modest. To achieve 10,000 connections per neuron (comparable to a mammalian brain), only a few devices would fit within a cm$^2$ (given the present model), and we are left in awe of the massive interconnectivity and scaling achieved by the bottom-up nanofabrication of biological organisms.
	
	While the scaling to 10,000 connections per neuron is formidable, the range of $N_{\mathrm{conn}} = 100-1,000$ is promising and technologically consequential. As is the case for scaling CMOS neuromorphic platforms, utilization of die tiling \cite{mear2014} will play a crucial role for this technology. For this purpose, the proposed SOEN platform is in an excellent position. Die can be tiled in 2D with several types of connectivity to adjacent die including electrical, single-flux quantum, and photonic communication over inter-die bridge waveguides. Additionally, tiling in the third dimension is possible with the usual bump bonding approach for electrical connectivity as well as with free-space optical signals sent from one chip using vertical grating couplers and received by a chip above or below using SNSPD arrays \cite{alve2015}. Information over such links can be encoded temporally, spatially, or in frequency with forgiving alignment tolerances. From Fig. \ref{fig:app_cnn_2}(b) we find that 700 neurons with 700 connections per neuron can fit on a 1 cm $\times$ 1 cm die if 10 waveguiding planes are utilized. We envision tiling a 100 $\times$ 100 array of these die in a sheet to build a system of $7\times 10^6$ neurons. This sheet will be $\approx 1$ mm thick, so to form a cube one meter on a side we envision stacking 1,000 vertical sheets. The system would then comprise $10^7$ die and $7\times10^{10}$ neurons, or roughly 10\% the number contained in the human brain. 
	
	To achieve such a system, we envision 1 m $\times$ 1 m sheets of 100 $\times$ 100 die mounted in trays with in-plane fiber optic connections leaving from the perimeter of the trays and out-of-plane free-space grating-to-SNSPD interconnects, thus enabling the trays to slide laterally. Achieving inter-sheet connectivity without physical bonds will enable access to die within the volume of the cube for diagnostics, repair, and local iteration and evolution. Massive interconnectivity between neurons on different die could be accomplished using such grating interconnects \cite{Fattal2010,Kildishev2013,Yu2014,Doylend2011,Sun2013}.
	
	Of greater importance than the size of highly scaled systems is the power consumption. We again consider a system of SPONs with 700 connections each. Such a device will consume $2\times10^{-17}$ J/synapse event, and with 700 connections, each firing event consists of 700 synapse events. Information processing in neuromorphic systems requires sparse event rates, so for the SOEN hardware wherein 20 MHz is achievable based on device limitations, 20 kHz represents a sparse rate. Note that this is a factor of $2\times10^4$\textendash$2\times10^5$ faster than biological event rates and a factor of 1,000 faster than the CMOS demonstration which achieved 26 pJ/synapse event and was limited by time-multiplexing \cite{mear2014}. For the system under consideration, we have $7\times10^9$ processing units which we consider to be firing at this rate with this energy per firing event, giving a total power consumption of 2 W. This equates to $5\times10^{16}$ synapse events per second per watt. To be fair, our system must be kept around 2 K, so an additional 1 kW/W must be supplied as cooling power, as discussed in Sec. \ref{sec:energy}. While this does not affect the power density (which ultimately limits scaling), and this 2 kW is minuscule compared to the tens of MW of a modern supercomputer, if we include this additional power in the calculation we find that we achieve $5\times10^{13}$ synapse events per second per watt.
	
	To put this in perspective, the human brain uses 20 W, but by analogy to the inclusion of the cooling power in the above calculation, one must include the human's total power of 100 W which is necessary to sustain the brain's operational state. The brain has roughly $10^{11}$ neurons with roughly $7\times10^3$ synapses per neuron firing between 0.1\textendash 1 Hz \cite{le2003,sqbe2008,lase2010,ai2015}. For the purposes of this calculation, we generously assume the rate is 1 Hz. This equates to $7\times10^{12}$ synapse events per second per watt. Even with the 1 kW/W cooling power of the cryostat, we find that the number of synapse events per second per watt of the SOEN system exceeds that of the brain by an order of magnitude. 
	
	Importantly, because signaling occurs predominantly in the optical domain, firing events can be directly imaged with a camera. For massively scaled systems, this becomes a powerful metrological tool. Such a measurement technique can be used to monitor device and system performance across spatial and temporal scales in a manner analogous to functional magnetic resonance imaging of biological organisms. 

	To close this discussion of scaling, we address the cryogenic requirements of a 1 m$^3$ SOEN system. We seek a $^4$He sorption refrigerator capable of cooling a 1 m$^3$ volume to 2 K with 2 W of cooling power. While this would be a relatively large cryostat, it is certainly well within the realm of possibility. No new physical principles of operation will need to be developed; it is simply a question of scaling up existing $^4$He cryogenic systems. Additionally, if suitable SNSPD materials can be found which operate at 4K with high yield, 2 W of cooling power is straightforward to achieve. We are of the opinion that with the advancement of single-flux-quantum processors, superconducting qubit devices, and SOENs, large-scale cryogenic technology will advance significantly in the coming years. Presently, many conversations in advanced computing debate whether the technology which proves victorious will operate within a cryostat or at room temperature. We speculate that a supercomputer of the future will leverage optoelectronic devices on various material platforms to employ quantum principles, neuromorphic principles, and digital logic principles across various temperature stages. The device designer is faced with the task of optimizing hardware performance at each temperature stage, and the architect is liberated to dream, with von Neumann, far beyond the architecture that now bears his name.  
	
		
	\section{\label{sec:discussion}Discussion and Outlook}
	We have explained the proposed devices and their functions, analyzed their performance, and considered their scaling. Here we consider possibilities for utilization of this platform for novel neuromorphic applications.
	
	\subsection{\label{sec:light}Advantages of optoelectronic neural networks}

	The unparalleled performance of the brain emerges from the enormous number of connections between neurons and the numerous complex signaling mechanisms available to the neurons. Optical signaling has an advantage over electronics in terms of the ability to route non-interacting signals in three dimensions without wiring parasitics. These strengths have been recognized for many years, and early implementations utilized reconfigurable holographic gratings \cite{Caulfield1989, Psaltis1990} for forming connections between optoelectronic neurons \cite{Psaltis1989}.  Even in a planar interconnect configuration, such as integrated waveguides, the non-interacting nature of photons enables signals from an arbitrary number of SPONs to be received simultaneously, thereby rendering time-multiplexing unnecessary. On an electronic platform, a shared interconnect can transmit only a single voltage pulse within a time window, and this limits both the number of connections between neurons and the firing rate of each neuron. Wiring parasitics also limit the number of electrical connections.
	
	Other approaches that leverage phenomena unique to optics for neuromorphic computing \cite{coge2011,Woods2012,nash2013,tana20142,tana2014,shna2016} have employed optical devices such as lasers and integrated microresonators. Laser cavities with strong light-matter interaction can be leveraged to realize complex nonlinear dynamics which can emulate the behavior of neurons \cite{coge2011,nash2013,shna2016}. The frequency selectivity of integrated ring resonators can be used to achieve synaptic weights \cite{tana20142}. Optical neural networks \cite{Mos1997,tana2014} and spiking neurons \cite{Kravtsov2011,Coomans2012, Hurtado2012, VanVaerenbergh2012,nash2013} based on these effects have been proposed and demonstrated. Optical reservoir computing has also recently been demonstrated \cite{Larger2012, Paquot2012, Vandoorne2014} as another way in which inherently optical phenomena can be leveraged for advanced computing. The distinction of the proposed SOEN platform is that it operates in the few-photon regime with compact, energy-efficient components, enabling a large degree of scalability. Thus, at present many electronic and photonic technologies appear promising for neuromorphic computing, and the most suitable hardware platform is likely to depend on the application.			
	
	\subsection{\label{sec:vc}The visual cortex}
	While we have described in detail in Sec. \ref{sec:cnn} how a simple neural network (the MLP) could be built with SPONs, the potential of the SOEN platform for more complex systems should not be overlooked. The visual cortex is the most thoroughly studied region of the mammalian brain \cite{laus2011}, yet there is still a great deal to be understood about information encoding from the retina through the thalamus and on to the visual cortex. A non-biological experimental testbed is highly desirable to explore hypotheses \cite{ri2009,paha2014}. Biologically realistic supercomputer simulations of the brain can only simulate a small fraction of the brain cells in a small mammal at significantly reduced speed \cite{Markram2006,Liu2010d}. The massive parallelism enabled by a scalable, biologically realistic hardware implementation of the many thousands of neurons involved in the visual system can provide more quick and efficient simulations \cite{Liu2010d,Poon2011,Hasler2013} which may give further insight into the visual system, while also offering potential for image processing applications.
	
	We are proposing a hardware platform with the potential for a built-in retina, manifest as integrated SNSPDs, which can be used in pixel arrays \cite{alve2015} for monolithic image acquisition and analysis. 
	\begin{figure*} 
		\centerline{\includegraphics[width=15.0cm]{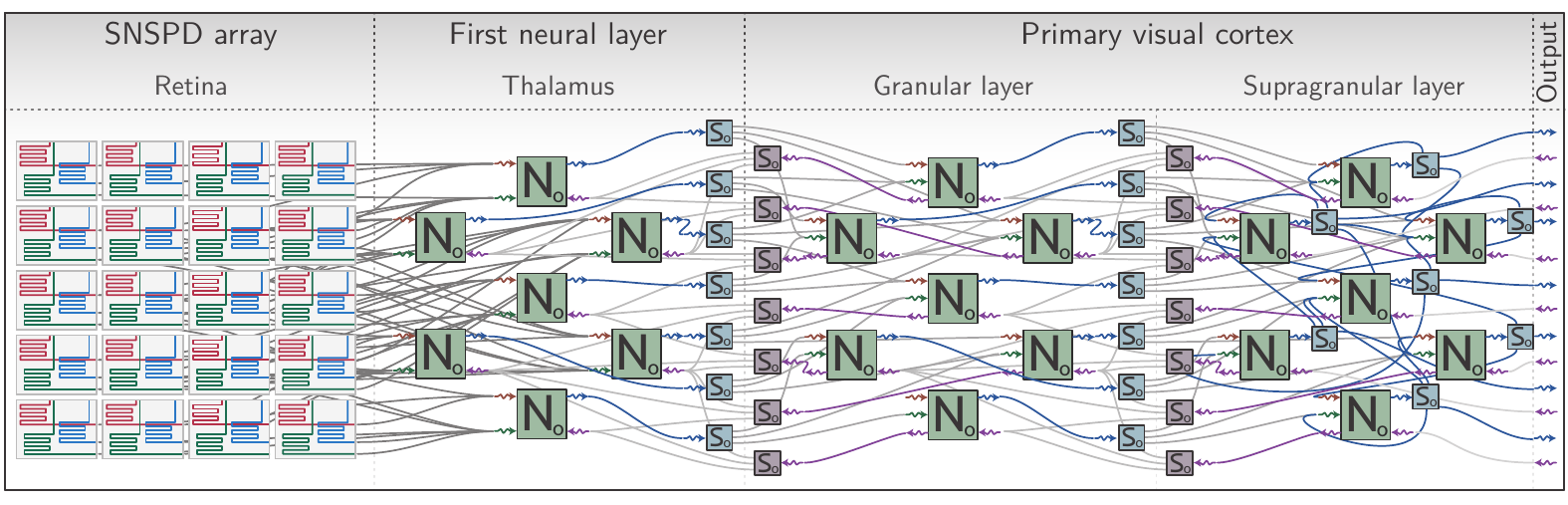}}
		\caption{\label{fig:app_vc}Schematic of a SOEN model of the mammalian visual cortex.}
	\end{figure*}
	In Fig. \ref{fig:app_vc} we show a schematic of how such an SNSPD array can be integrated with a multilayer neural network to emulate the visual system. To illustrate the key points of such an experimental system, we break the visual system into three parts: the retina, the thalamus, and the primary visual cortex. In biological systems, the primary visual cortex is highly sophisticated, being organized into six layers each with their own sublayers \cite{laus2011}. For the purpose at hand, we treat the primary visual cortex as being composed of two layers, referred to as the granular layer and the supragranular layer.
	
	At the left of Fig. \ref{fig:app_vc}, the SNSPD array receives light from the environment and converts it to signals to be sent to the first layer of neurons in the thalamus, in direct analogy with a biological retina. Much like the cones in one's eye, the pixels of the SNSPD array could be designed to be more sensitive to particular frequencies simply by varying the thickness of an anti-reflection coating locally above each pixel.
	
	From the retina, a small number of pixels project to each neuron in the thalamus without a large amount of branching. Similarly, the neurons of the thalamus project to the first layer of the visual cortex with minimal branching. Importantly, some of these connections are inhibitory, and some are excitatory. While inhibitory connections are known to play a central role in information encoding in the visual system, the full scope of that role remains the subject of investigation. The biologically realistic mechanism for implementing inhibitory connections, as illustrated in Fig. \ref{fig:circuits_variants}(c) is of great utility in using SOENs to study information encoding in the visual system. In the thalamus, there is little if any recurrence, meaning the neurons in that layer project forward, but do not form synapses on each other. The thalamic neurons do, however, receive feedback from the granular layer of the visual cortex. The ability to straightforwardly implement feedback with SOENs, as illustrated in Fig. \ref{fig:circuits_variants}(e), is another feature of great utility in using SOENs to model the visual system.
	
	The granular layer receives feedforward signals from the thalamus, projects feedforward signals to the supragranular layer, and receives feedback from the supragranular layer. While still only minimally recurrent, neurons in the granular layer branch more heavily to form a larger number of connections across more neurons in the supragranular layer. The supragranular layer projects its output to other regions of cortex and is also heavily recurrent. At the right of Fig. \ref{fig:app_vc}, we have shown the neurons in the supragranular layer making connections with other neurons within the layer.
	
	For an initial SOEN visual system, we envision implementing the retina and thalamus on a single die, with a separate chip of 700  neurons being employed for the granular layer, and a third chip of 700 mutually interacting neurons representing the supragranular layer. This experimental testbed may offer insight into outstanding  questions such as how and why concentric circular patterns of retinal response are mapped to bars for processing in the visual cortex. With a simple system like that illustrated in Fig. \ref{fig:app_vc}, it will be possible to conduct experiments related to object recognition, edge detection, the perception of motion and spatial frequency, as well as many other subjects in contemporary visual system research.
	
	\subsection{\label{sec:highPerformance}High-performance application spaces}
	One strength of neuromorphic systems is their ability to find trends and extract features from large and noisy data sets, thereby reducing the dimensionality of those data sets \cite{hisa2006}. They can learn over time based on the temporal evolution of the data under consideration. Several societal challenges require this type of analysis of large numbers of complex, interacting units\textemdash exactly the type of system for which neuromorphic computing excels. These applications include monitoring of markets, internet traffic metrology, detection of hacking attacks, modeling of climate systems, and phenotypic prediction from genomic data. For these applications, supercomputers at the limit of what is possible with CMOS implementations of the von Neumann architecture are presently in use. Yet greater performance is still required. For the most demanding computational tasks of this class, massively scaled systems employing parallel computation in a neuromorphic architecture are likely to play a central role. It is for these applications that we envision the SOEN platform making the largest impact.
	
	Another likely solution to the current bottlenecks facing supercomputers is superconducting electronics. In particular, Josephson-junction processors with single flux quantum logic are poised for use in the next generation of supercomputers. These processors can provide an improvement over CMOS in speed by roughly a factor of 100 with extremely high energy efficiency.  Our proposed platform will integrate well into such supercomputers, offering neuromorphic capability to von Neumann implementations  \cite{lise1991} and additional degrees of freedom to neuromorphic Josephson-junction systems \cite{hias2007,crsc2010,segu2014}, which are purely electronic. In addition, the SOEN platform may offer a means to transduce single-flux-quantum pulses to the optical domain, for interconnects between chips and with the outside world (cryostat I/O) via photonic signaling.

	\subsection{\label{sec:summary}Summary}

	We have described a hardware platform combining superconducting single-photon detectors and electronics with semiconducting faint photon sources to operate as a massively interconnected information processing system. The SOEN platform consists of neurons that exhibit complex signaling and efficient access to photonic degrees of freedom such as frequency, polarization, mode index, intensity, and coherence, in analogy to the complex signaling mechanisms in the brain. The proposed networks of connections, based on reconfigurable waveguides, offer advantages over electronic connections in terms of speed, connectivity, and energy efficiency.
	
	In the present paper, we have argued that through the use of networks of neurons consisting of semiconductor LEDs, superconducting nanowire single photon detectors, and reconfigurable optical waveguides, we can build advanced computing systems. Such networks could achieve states of enormous entropy through massive interconnectivity and the interaction of multiple physical degrees of freedom. We have further shown that the integrate-and-fire operation of superconducting optoelectronic neurons can be used for spike-encoding information. Such spike-encoded information is highly advantageous for high-bandwidth information processing with temporal information encoding and resilience to noise. These concepts have recently been placed on a solid theoretical foundation \cite{rapo2016,pola2016,lite2016,haah2015}, so we should not be surprised to find that the brain's computing mechanisms employ all of these concepts. The fundamental principles of information theory which enable reasoning, decision, innovation, and consciousness are currently incompletely understood. To date we know of only one computing platform which can accomplish these tasks: the biologically evolved neural system. We do not appear to be close to a complete understanding of the information theory describing such a complex system. Yet by exploring alternative physical systems with comparable complexity we stand to learn a great deal about the fundamentals of information science.

	\section{Acknowledgements}
	We wish to thank Prof. Ryan Tung, Dr. Amanda Casale, Dr. Aaron Hagerstrom, Dr. Ann Hermundstad, Dr. Matthew Phillips, Prof. Gert Cauwenberghs, and Ryan Uhlenbrock for fruitful discussions. \\
	\vspace{0.5em} \\
	This is a contribution of NIST, an agency of the US government, not subject to copyright.
		
	\newpage
	
	\appendix

	\section{\label{apx:PNDAppendix}Threshold condition for the PND array}
	Here we derive the expression of Eq. \ref{eq:nc}. We begin by defining all the quantities of interest. 
	\begin{center}
	\vspace{1em}
	\begin{tabular}{ | r | l | }
	\hline
	\parbox[c][2.3em][c]{5em}{ \vspace{0.5em} \raggedleft \textbf{Symbol} \hspace{0.5em} } & \parbox[c][2.3em][c]{10em}{ \vspace{0.5em} \raggedright \hspace{0.5em} \textbf{Meaning} } \\
	\hline
	\parbox[c][3em][c]{4em}{ \vspace{1em} \raggedleft $N_{\mathrm{nw}}$ \hspace{0.5em} } & \hspace{0.5em} \parbox[c][4em][c]{0.65\linewidth}{\vspace{0.5em} \raggedright Number of nanowires in the PND array } \\
	\hline
	\parbox[c][3em][c]{4em}{ \vspace{1em} \raggedleft $n^{\mathrm{abs}}$ \hspace{0.5em} } & \hspace{0.5em} \parbox[c][4em][c]{0.65\linewidth}{ \vspace{0.5em} \raggedright Number of nanowires driven normal by photons }  \\
	\hline
	\parbox[c][3em][c]{4em}{ \vspace{1em} \raggedleft $n^{\mathrm{abs}}_\mathrm{c}$ \hspace{0.5em} } & \hspace{0.5em} \parbox[c][4em][c]{0.65\linewidth}{ \vspace{0.5em} \raggedright Critical number of nanowires driven normal } \\
	\hline
	\parbox[c][3em][c]{4em}{ \vspace{1em} \raggedleft $I_{\mathrm{b}}$ \hspace{0.5em} } & \hspace{0.5em} \parbox[c][4em][c]{0.65\linewidth}{ \vspace{0.5em} \raggedright Bias current through the entire array } \\
	\hline
	\parbox[c][3em][c]{4em}{ \vspace{1em} \raggedleft $i$ \hspace{0.5em} } & 
	\hspace{0.5em} \parbox[c][4em][c]{0.65\linewidth}{ \vspace{0.5em} \raggedright Current through a single wire of the array } \\
	\hline
	\parbox[c][3em][c]{4em}{ \vspace{1em} \raggedleft $i_{\mathrm{c}}$ \hspace{0.5em} } & \hspace{0.5em} \parbox[c][4em][c]{0.65\linewidth}{ \vspace{0.5em} \raggedright Critical current of a single wire } \\
	\hline
	\end{tabular}
	\vspace{1em}
	\end{center}
	
	In the steady state, before any photons have been absorbed, $n^{\mathrm{abs}} = 0$, and $i = \frac{I_{\mathrm{b}}}{N_{\mathrm{nw}}}$. Upon absorption of a single photon, $n^{\mathrm{abs}} = 1$, and $i = \frac{I_{\mathrm{b}}}{N_{\mathrm{nw}}-1}$. In the general case that $n$ nanowires have been driven normal by photons, $n^{\mathrm{abs}} = n$, and $i = \frac{I_{\mathrm{b}}}{N_{\mathrm{nw}}-n}$. 
	
	The condition for $n^{\mathrm{abs}}_c$ is $i = i_{\mathrm{c}} = \frac{I_{\mathrm{b}}}{N_{\mathrm{nw}}-n^{\mathrm{abs}}_c}$.	Rearranging gives $n^{\mathrm{abs}}_c = N_{\mathrm{nw}}-\frac{I_{\mathrm{b}}}{i_{\mathrm{c}}}$.

	\section{\label{apx:SNSPDAppendix}Integration of superconducting nanowire detectors}
\begin{figure} 
		\centerline{\includegraphics[width=7.0cm]{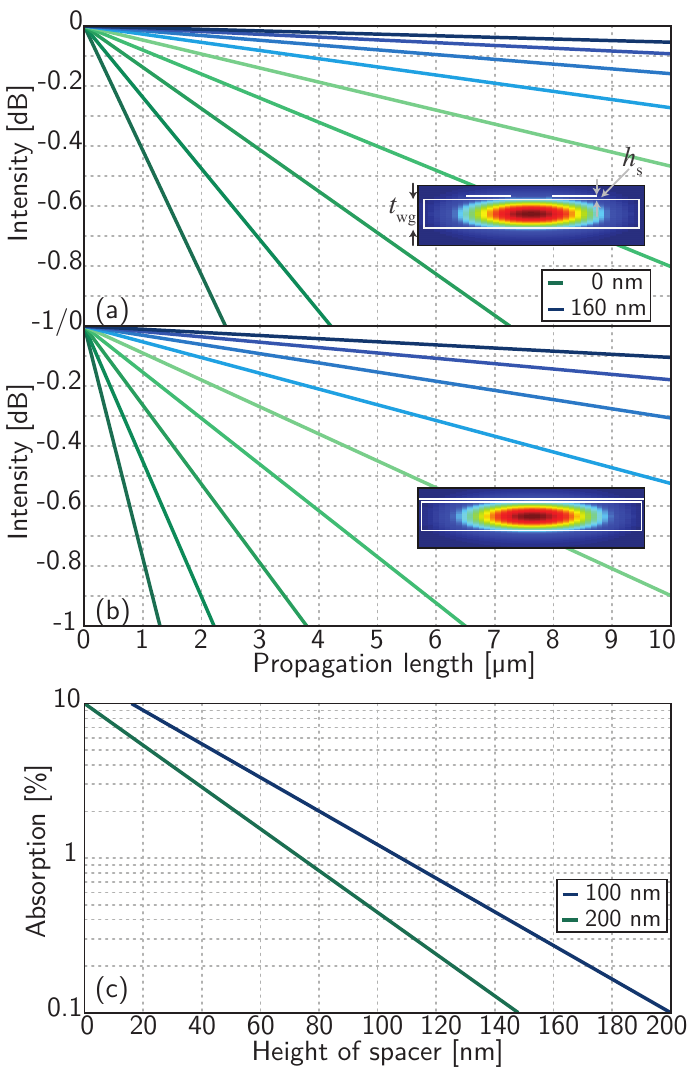}}
		\caption{\label{fig:absorption}Absorption of light propagating in a waveguide with SNSPD on top in (a) parallel and (b) perpendicular configurations, for different spacer heights between the SNSPD and waveguide. (c) Absorption in waveguides of different thicknesses for different spacer heights.}
	\end{figure}

	To properly understand the behavior of the SNSPD receivers, we must analyze the optical absorption and statistical behavior of waveguide-integrated SNSPDs \cite{spga2011,pesc2012,feka2015,namo2015,saga2015,scgu2016,shbu2016}. We first calculate the attenuation of light as a function of propagation length for 200 nm thick waveguides ($t_\mathrm{wg}$) in the asymptotic slab regime. The waveguide refractive index is 3.52, the cladding index is 1.46, and our calculations are at a wavelength of 1220 nm. The nanowire is assumed to be 4 nm thick, 300 nm wide with 50 \% fill factor and $n = 3.25 + 2.19i$. In Fig. \ref{fig:absorption} we show the results for the common out-and-back configuration [light propagating parallel to nanowire, Fig. \ref{fig:absorption}(a)] and the slab configuration [light propagating perpendicular to nanowire, Fig. \ref{fig:absorption}(b)]. In each case, the various traces are for different spacer thicknesses, ($h_s$, refractive index 1.46) between the waveguide and nanowire, ranging from zero to 160 nm in steps of 20 nm. The modal distribution is shown in the inset. The data in Figs. \ref{fig:absorption}(a) and (b) is fractal in nature, so an increase of the $x$-axis by one decade is accompanied by an increase in the $y$-axis by a decade (on the dB scale). From these plots one can see that for both the parallel and perpendicular configuration, a wide range of attenuation coefficients can be achieved. 
	
	In Fig. \ref{fig:absorption}(c) we show the probability of absorption after a single pass by a nanowire as a function of spacer thickness for waveguides with 100 nm and 200 nm thickness, illustrating another degree of freedom for tuning the absorption. It is important to be able to engineer the statistical distribution of absorption across the SNSPD receiver. For the case of the PND, each SNSPD should absorb an average of one photon each, as an additional photon absorption in the same SNSPD will not contribute to the spike event. For the case of the SND, the requirement is less stringent, but one would still like to spatially distribute absorption events so that hotspots do not overlap until a certain (large) number of photons has been absorbed. 
	
\begin{figure} 
		\centerline{\includegraphics[width=7.0cm]{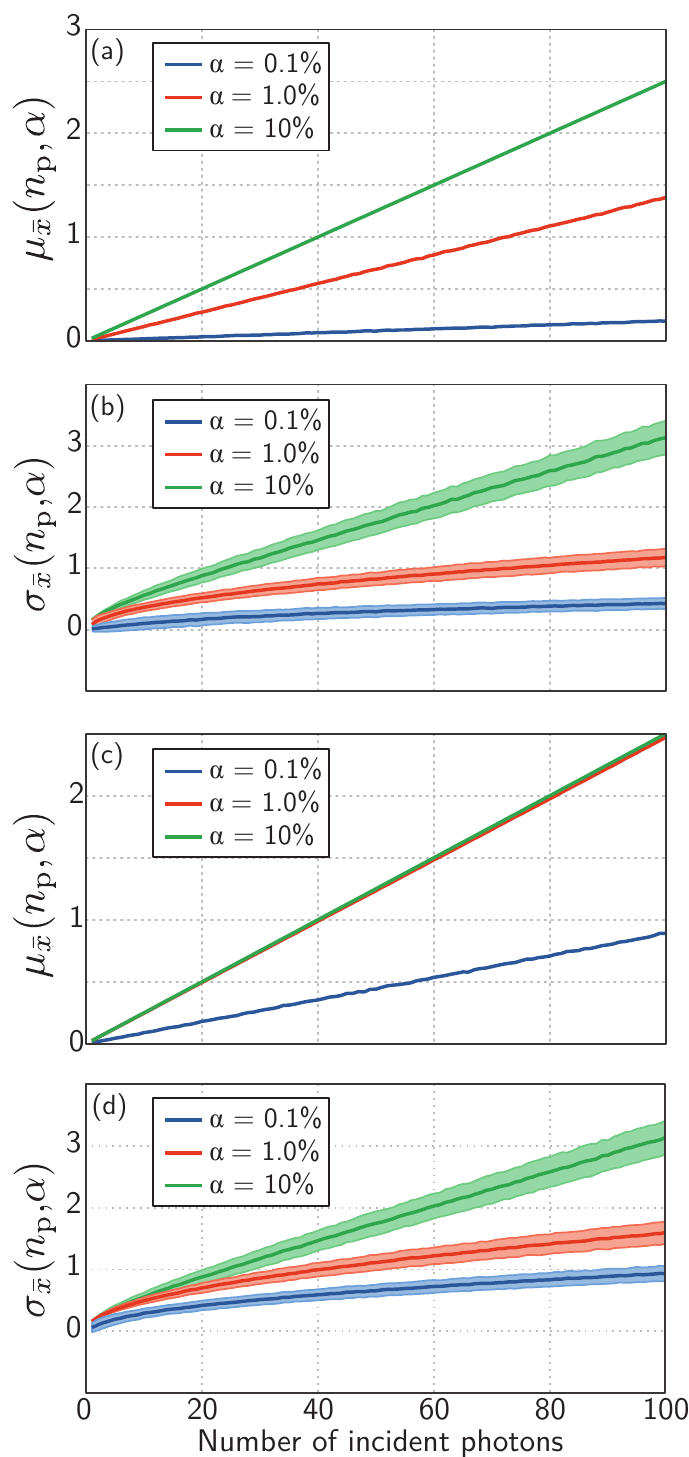}}
		\caption{\label{fig:absorptionStatistics} Mean number (a) and standard deviation (b) of absorbed photons versus number of incident photons for neuron designs where light is directed past each nanowire once (single-pass). Mean number (c) and standard deviation (d) of absorbed photons versus number of incident photons for neuron designs where light is directed past each nanowire ten times.}
	\end{figure}

	To address the design requirements of the PND, we consider the absorption statistics as calculated via Monte Carlo simulations. We perform 1,000 trials each for different photon numbers incident on a PND with 40 SNSPDs. Figure \ref{fig:absorptionStatistics}(a) shows the mean number of photons absorbed (out of 1000 trials) in the PND as a function of the number of incident photons for different absorption probabilities, in the case where only a single pass by each nanowire occurs. This may be achieved with a design like that of Fig. \ref{fig:stingray}. For each of the 1,000 simulations, the arithmetic mean of the number of photons absorbed per nanowire was calculated for each value of incident photon number as

	\begin{equation}
	\label{eq:mean}
	\mu_{x}(n_{\mathrm{p}},\alpha) = \frac{1}{N_{\mathrm{nw}}}\sum_{i = 1}^{N_{\mathrm{nw}}}x_i,
	\end{equation}
	where $x_i$ is the number of photons absorbed in the $i$th nanowire. From these values, the mean number of photons absorbed per nanowire $\mu_{\bar{x}}$ was then calculated as the mean of the means (grand mean) in Eq. \ref{eq:mean}.
	
	One would like to engineer the absorption probability in the PND such that the mean number of absorbed photons per nanowire per pulse and the standard deviation of this number are both less than or equal to one. In Fig. \ref{fig:absorptionStatistics}(b) we show the standard deviation data for the single-pass case. For each of the thousand trials, the standard deviation of the number of absorbed photons was calculated as 
	\begin{equation}
	\label{eq:sigma}
	\sigma_x(n_{\mathrm{p}},\alpha) = \sqrt{ \frac{1}{N_{\mathrm{nw}}} \sum_{i = 1}^{N_{\mathrm{nw}}} (x_i-\mu_x)^2 },
	\end{equation}
	where $\mu_x$ is given by Eq. \ref{eq:mean}. The mean of these standard deviations over the 1,000 Monte Carlo trials ($\sigma_{\bar{x}}$) was calculated, as was the standard deviation of the standard deviations. The center trace of each curve in Figs. \ref{fig:absorptionStatistics}(b) is $\sigma_{\bar{x}}$ for a given value of $\alpha$, and the width of the trace is calculated by adding and subtracting the standard deviation of the standard deviations. The standard deviation with $\alpha = 10$\% is roughly three photons. Thus, such large absorption is undesirable for this purpose as the initial wires tend to absorb more than a single photon, and the latter wires absorb zero photons. For the one-pass case, 1\% absorption appears to be close to ideal. The mean number of absorbed photons is close to one, as is the standard deviation. The standard deviation for $\alpha = 0.1$\% is even lower, yet the mean number of absorbed photons is only $\approx 0.2$. Therefore, many photons are passing through the array without being absorbed. 

	In Fig. \ref{fig:absorptionStatistics}(c) and (d) we show results for the case where 10 passes by each nanowire occur, as may be achieved with the spiderweb neuron design of Fig. \ref{fig:spiderweb_2}. 
	For the case of 10 passes, $\alpha = 0.1$\% performs much better, although all photons are still not absorbed. 

	Consider the case where 40 photons are incident. We would like all 40 of these photons to be absorbed by the 40 nanowires of the array, and therefore we would like $\mu_{\bar{x}}$ to be near unity. In Fig. \ref{fig:absorptionStatistics}(c) we see that we achieve this for both $\alpha = $1\% and 10\%, yet in the case of $\alpha = $10\% all photons are absorbed on the first pass [as seen in Fig. \ref{fig:absorptionStatistics}(a)], so the mode of the distribution is greater than one, and the standard deviation is larger than desired. By comparing the standard deviations for the $\alpha = 1$\% and $\alpha = 0.1$\% cases in Fig. \ref{fig:absorptionStatistics}(d), we find that $\alpha = 0.1$\% gives a more desirable spread of absorption events (smaller standard deviation). From this analysis we find that for the PND receiver array, it is desirable to operate with low $\alpha$ a high number of passes.  
	
	\section{\label{apx:time}Integration time and refractory period}
	The integration time of a SPON is the time from the absorption of a photon until the receiver no longer has a memory of that absorption event. The behavior of integrate-and-fire devices with integration times less than infinity are referred to as leaky integrate-and-fire neurons. In the context of SPON devices, in the most basic case, this is determined by the hotspot relaxation time of the superconductor, which depends on the material quasiparticle dynamics which are governed by the electron-phonon coupling and the thermal conduction to the substrate. This is a material-dependent quantity, and can be as fast as 200 ps in NbN \cite{mast2016}. In WSi it is closer to 1 ns \cite{mast2016}, and there may be materials for which it is even slower. Additionally, the bias current has been shown to have a significant effect on quasiparticle recombination time \cite{mast2016}. Therefore, choice of superconducting material and substrate may be leveraged to tune the integration time to a desired value in hardware, and the bias current may be used to modify it dynamically.
	
	Further, the PND circuit shown in Fig. \ref{fig:circuits_PND} can be modified so that each wire in the PND array is in parallel with a small shunt resistor. In this configuration, the $L/R$ time constant of each receiving wire can be chosen to set the integration time. In this case, the hotspot relaxation time represents a lower limit on the integration time, but the integration time can be extended to very long times relative to other time scales of the system simply by adjusting the $L/R$ value. 
	
\begin{figure} 
			\centerline{\includegraphics[width=7.0cm]{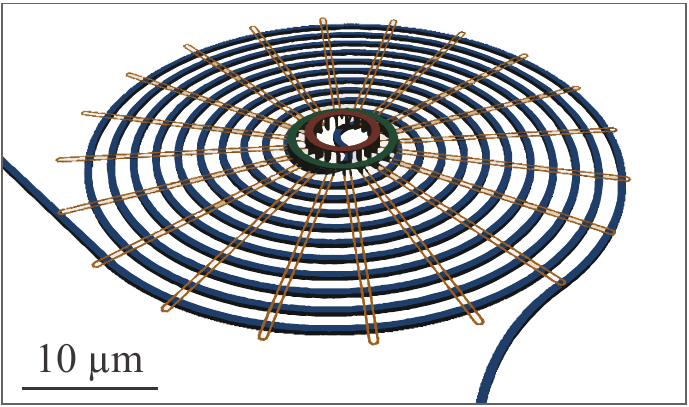}}
			\caption{\label{fig:spiderweb_3} Flux-dissipating version of the spiderweb neuron.}
	\end{figure}

\begin{figure} 
		\centerline{\includegraphics[width=7.0cm]{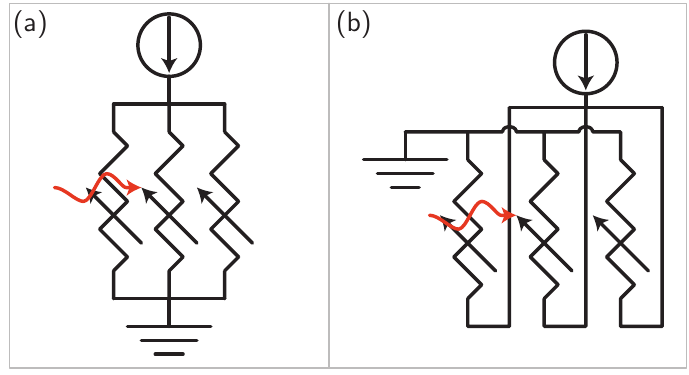}}
		\caption{\label{fig:circuits_PND_flux} (a) Flux-trapping PND circuit. (b) An alternative PND design which avoids flux trapping.}
	\end{figure}

	Recent studies \cite{cahe2013,cahe2015} reveal that non-uniform current distribution in the PND as drawn in Fig. \ref{fig:circuits_PND}(a) is problematic for number-resolving photon detection. To avoid this, the cylindrically symmetric nanowire arrays of Fig. \ref{fig:spiderweb_1} and \ref{fig:spiderweb_3} are proposed. In this geometry, no nanowire occupies an edge, so supercurrent is evenly distributed after each firing event. Also shown in Ref. \cite{cahe2015} is the fact that a PND can trap flux after a photon absorption event. To utilize this to extend the integration time to infinity, the geometry of Fig. \ref{fig:spiderweb_1} is proposed. If one wishes to dissipate flux to reduce the integration time, the topological variant of Fig. \ref{fig:spiderweb_3} is proposed. The differing circuit designs of these two devices are shown in Fig. \ref{fig:circuits_PND_flux}. In the flux-dissipating configuration shown in Figs. \ref{fig:spiderweb_3} and \ref{fig:circuits_PND_flux}(b), flux-trapping superconducting loops are avoided, and all locations where hotspots can be created are on a boundary with the normal environment. Therefore, vortices created during absorption events are not trapped.
			
	In addition to utilizing flux trapping to extend the integration time, one may design the $L/R$ time constant to achieve desired performance. In the original studies of the PND for number-resolving detection \cite{dima2008,mabi2009}, each nanowire was in series with a resistor, while the proposal for the SND for number-resolving detection \cite{jafi2012} utilized the dual circuit wherein each series nanowire element is in parallel with a resistor. While the series resistors of the PND are of limited utility in this application due to their addition of power consumption in the steady state, the parallel resistors for the SND can be employed with no steady-state power penalty, and the choice of nanowire element inductance and parallel resistance can thereby be used to engineer the desired integration time.
	
	We note that in biological systems, the integration time is set by the $RC$ time constant of the membrane, and is typically $\approx 1$ ms, or $\approx 10^{-4}$\textendash $10^{-5}$ the firing period. Taking the 1 ns quasiparticle lifetime as the integration time, this would correspond to operating the system with 10-100 kHZ event rates, a range that is straightforward to achieve.
	
	The refractory period of a neuron refers to the time following a firing event during which the neuron cannot fire again. For a standard SNSPD, this dead time is governed by the $L/R$ time constant of the series inductance of the SNSPD and the resistance across which the voltage pulse is being measured. In the case of WSi this is usually 50 ns \cite{yake2007}. This resistance is usually 50 $\Omega$, but in the present case it is the impedance of the LED, which will be several k$\Omega$, giving a shorter refractory period. If an application requires a longer refractory period, an additional series inductance can be added to achieve the desired delay. We note that in some SNSPD material systems, the $L/R$ time constant must be chosen sufficiently large to avoid latching, while in the present application the feedback circuit of Fig. \ref{fig:circuits_variants}(e) can also be utilized to avoid latching and control the refractory period. 
		
	\section{\label{apx:LEDAppendix}$p-n$ junction model of the light-emitting diode}
	To model the performance of the emitters, we worked with an analytical model of a $p-n$ junction \cite{stba2006}. Within this model, the current-voltage relationship for the junction is given by
	\begin{equation}
	\label{eq:pn}
	I_{\mathrm{pn}}(V) = eA\bigg( \sqrt{\frac{D_p}{\tau_p}}p_n + \sqrt{\frac{D_n}{\tau_n}}n_p \bigg)\big( \mathrm{e}^{eV/k_{\mathrm{B}}T} -1 \big).
	\end{equation}
	In Eq. \ref{eq:pn}, the electron and hole diffusion coefficients are given by $D_n=\mu_n(kT/e)$ and $D_p=\mu_p(kT/e)$, where $\mu_n$ ($\mu_p$) is the mobility of electrons (holes). The electron and hole lifetimes are given by $\tau_n$ and $\tau_p$, respectively, which we take to be 40 ns. $n_p$ is the concentration of electrons on the $p$-doped side of the junction, and $p_n$ is the concentration of holes on the $n$-doped side of the junction. To achieve low-temperature operation, we assume degenerate doping, and therefore a low mobility is to be expected. We have used a value of $100 \mathrm{cm}^2/(\mathrm{V}\cdot \mathrm{s})$ \cite{arha1982} for both electron and hole mobilities. Because this value will be limited by ionized impurity scattering, it is likely to change little as the temperature is decreased to 1 K.
	
	From the electronic current, we calculate the photonic current as
	\begin{equation}
	\label{eq:photonCurrent}
	I_{\nu}(V) = \eta \frac{I_{\mathrm{pn}}(V)}{e}.
	\end{equation}		
	This model for the current through the diode is derived for an abrupt $p-n$ junction, yet for the waveguide-integrated LED one would employ a $p-i-n$ junction. Also, the present model breaks down at low temperature. We have used $T = 300$K in Eq. \ref{eq:pn}, because our measurements inform us that in the degenerate doping regime, behavior is relatively constant to low temperature. Therefore, we use this model only as an approximation, and more thorough numerical and experimental investigation of the devices to be used in the platform will be the subject of future investigation. With this in mind, we approximate the capacitance of the junction using a simple parallel plate model where the capacitance is given by $C = \epsilon A/d$ where $\epsilon$ is the material permittivity, $A$ is the capacitor area, and $d$ is the distance between the plates. We assume $\epsilon = 12 \epsilon_0$, $A = 10 \mu\mathrm{m} \times 100$ nm, and $d = 300 nm$. The energy associated with charging this capacitor is then calculated as $E_c = 1/2 \mathrm{CV}^2$. We note that for all values of photon number generated by the LEDs within this model, the applied voltage was below the built-in potential of the junction, so true forward-bias operation was not required. We anticipate that for the case of a $p-i-n$ junction, the voltages required to achieve the same number of photons will increase slightly, but this can easily be accommodated by utilizing nanowires with larger critical currents. 
	
	\section{\label{apx:wgAppendix}Waveguide design for the dendritic arbor}

\begin{figure} 
		\centerline{\includegraphics[width=7.0cm]{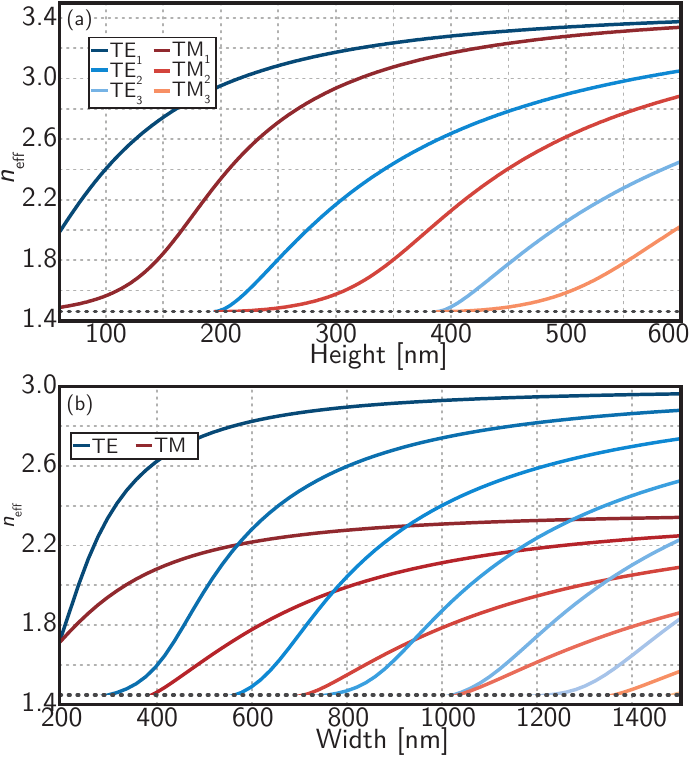}}
		\caption{\label{fig:nEff} Effective indices of refraction for various guided modes in a waveguiding layer with index of refraction $n = 3.52$ and cladding $n = 1.46$. (a) Slab mode calculations of both TE and TM modes for different film thicknesses showing different vertical mode orders. (b) TE and TM modes for different waveguide widths in a film of height 200 nm. The cladding index is shown as the dashed line in both (a) and (b).}
	\end{figure}

	 In Fig. \ref{fig:nEff}(a) we show effective indices at 1220 nm for slab thicknesses up to 600 nm to illustrate that many vertical modes can be present with high effective indices with only modest film thicknesses. We find that for $<200$ thick waveguides, only the first vertical order TE and TM modes are present. Therefore in Sec. \ref{sec:dendriticArbor}, we have assumed a waveguide height of 200 nm. For massive scaling even beyond that presented in Sec. \ref{sec:scaling}, it may be necessary to use multimode waveguides with higher vertical as well as lateral modes and both polarizations.
	
	Having selected 200 nm as our waveguiding layer thickness, we consider the lateral mode spectrum, as shown in Fig. \ref{fig:nEff}(b). Here we see that the second-lateral-order TE mode emerges above the cladding index around 350 nm; we choose this as the single-mode width for the dendritic arbor simulations. From Fig. \ref{fig:nEff}(b) we also find that a large number of higher-lateral-order modes are present with high effective index and modest waveguide width. For the dendritic arbor design presented in Fig. \ref{fig:spiderweb_2}(b) it is important that a compact, multimode waveguide be achievable. From this analysis we find that a waveguide with tens of modes can be achieved while still maintaining a compact bend radius.
	
\begin{figure} 
		\centerline{\includegraphics[width=7.0cm]{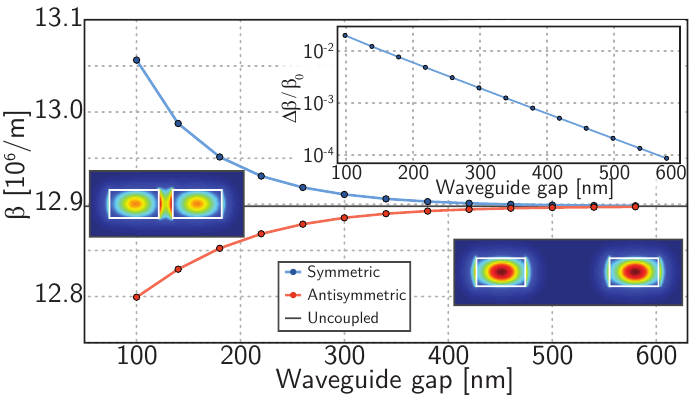}}
		\caption{\label{fig:wgCoupling} Supermode propagation constants for 200 nm thick, 350 nm wide waveguides with 3.52 core index and 1.46 cladding index at $\lambda = 1220$ nm. The inset shows the fractional splitting, and the mode profiles show the symmetric mode for gaps of 100 nm and 600 nm.}
	\end{figure}

	In addition to choosing the single-mode width, we also need to choose the minimum inter-waveguide gap that will avoid undesired coupling of modes in space. To do this, we calculate the supermode propagation constants as a function of waveguide gap, as shown in Fig. \ref{fig:wgCoupling}. We see the splitting between the symmetric and antisymmetric modes is quite large for a gap of 100 nm, but both modes converge to the uncoupled value for a gap of 600 nm. The fractional splitting, $\Delta\beta/\beta_0$, is shown in the inset. Here, $\Delta\beta$ is the difference between the propagation constants of the symmetric and antisymmetric supermodes, and $\beta_0$ is the uncoupled propagation constant. Based on this analysis, we choose 600 nm to be the inter-waveguide gap for the dendritic arbor design of Fig. \ref{fig:stingray} and the value used in the scaling analysis of Sec. \ref{sec:scaling}.	
	
	\section{\label{apx:scaling}Scaling}
	In reference to Sec. \ref{sec:scaling}, the length of an MLP layer is given by
	\begin{equation}
	\label{eq:length}
	L_l = (L_t+L_g+L_x)\frac{N_n}{N_{wg}}+2L_{wg}N_{wg}+L_n,
	\end{equation}
	where $L_t$ is the length of a single tap (or synapse), taken to be $10 \mu$m; $L_g$ is the length of a gap between two vertically running waveguides, taken to be $5 \mu$m, which is sufficiently wide to allow for undercut of the mechanically mobile synapses; $L_x$ is the length of a intralayer waveguide crossing, taken to be $3 \mu$m; $N_n$ is the number of neurons in an MLP layer [four in Fig. \ref{fig:app_cnn_1}(a)]; $N_{\mathrm{wg}}$ is the number of vertically stacked waveguide planes used for routing; $L_{wg}$ is the length of an interlayer coupler between two waveguiding layers, taken to be $10 \mu$m. $L_n$ is the length of a single neuron as shown in Fig. \ref{fig:stingray}. $L_n$ is determined predominantly by the number of inputs, and therefore is taken to be the interwaveguide gap, 600 nm $\times N_n$. The width of a single neuron is taken to be equal to its length, and within this model we are assuming each neuron in a given layer has a synapse connecting to each neuron of the next layer.  
	
	\section{\label{apx:info}Information}
	Application of Shannon's theory of communication \cite{sh1948} to neural systems enables the quantification of information processing capacity. The mutual information (in bits) between a neural system and a stimulus can be represented as \cite{daab2001}
	\begin{equation}
	\label{eq:info}
	I_{m}=\int\mathrm{d}s\int\mathrm{d}rP[s]P[r|s]\mathrm{log}_{2}\bigg(\frac{P[r|s]}{P[r]}\bigg).
	\end{equation}
	In Eq. \ref{eq:info}, $P[r]$ is the probability of spike rate $r$ occurring given a stimulus $s$, $P[s]$ is the probability of stimulus $s$ occurring from the set of all possible stimuli, and $P[r|s]$ is the conditional probability of response rate $r$ being evoked when the system is presented with stimulus $s$. With a neuromorphic computing platform, one would like to maximize the mutual information. Because $I_m$ within this model is calculated simply as a double integral over stimuli and response rates, we can maximize this quantity by increasing the limits of the integral. Because the proposed devices can operate at 20 MHz\textemdash and potentially up to 1 GHz by employing superconductors with faster thermal recovery\textemdash they can achieve response rates as well as receive stimulus across this entire bandwidth. The intrinsic speed of SPONs is greater than biological systems by a factor of $10^4$, and this affects both the stimulus and response bandwidths in the double integral.
	
	In addition to increasing the double integral by increasing the bandwidths, we can also maximize the bit depth. As discussed in Sec. \ref{sec:scaling}, signals can be discretized into roughly 11 bits. However, it is possible to increase this number further at the expense of size and efficiency.
	
	We have been discussing the $s$ and $r$ in Eq. \ref{eq:info} with the photonic input to the receiver array and photonic output pulse rate of the transmitter in mind, but the neuron of Fig. \ref{fig:N} can receive more stimulus and generate more output. For example, if one considers not only the photons incident upon the receiver as stimulus but also the current through the SNSPD, bit depth of the discernible stimuli increases further. 
	
	Equation \ref{eq:info} is derived by considering the difference between the entropy of a neuron's responses to a given stimulus and the noise entropy. As such, it is a measure of the information content at the device level and not at the system level. A full analysis of the information content of population-encoded information is beyond the scope of this work. At a minimum, we point out that the information content of a population grows with the size of that population. Therefore, the high-bandwidth of SPON devices, the ability to scale to units with large numbers of connections, and the ability to scale to systems with large numbers of units while maintaining a low power density points to the potential for complex systems with enormous information content. We note that these attributes are enabled by photonic signaling and superconducting electronics.		
	
	\bibliography{SOENs,SOENsMendeley}

\end{document}